\documentclass[11pt]{article}
\usepackage[preprint]{acl}
\usepackage[T1]{fontenc}
\usepackage[utf8]{inputenc}
\usepackage{lmodern}
\usepackage{microtype}
\usepackage{orcidlink}
\usepackage{amsmath}
\usepackage{amssymb}
\usepackage{booktabs}
\usepackage{tabularx}
\usepackage{array}
\usepackage{calc}
\usepackage{graphicx}
\usepackage{xcolor}
\usepackage{fancyvrb}
\usepackage{fvextra}
\usepackage{framed}
\usepackage[htt]{hyphenat}
\usepackage{ragged2e}
\usepackage{seqsplit}

\DeclareUnicodeCharacter{00A7}{\S}
\DeclareUnicodeCharacter{00D7}{\ensuremath{\times}}
\DeclareUnicodeCharacter{2013}{--}
\DeclareUnicodeCharacter{2019}{'}
\DeclareUnicodeCharacter{201C}{``}
\DeclareUnicodeCharacter{201D}{''}
\DeclareUnicodeCharacter{2192}{\ensuremath{\rightarrow}}
\DeclareUnicodeCharacter{2212}{\ensuremath{-}}
\DeclareUnicodeCharacter{2264}{\ensuremath{\leq}}
\DeclareUnicodeCharacter{2265}{\ensuremath{\geq}}

\definecolor{shadecolor}{RGB}{248,248,248}

\DefineVerbatimEnvironment{Highlighting}{Verbatim}%
  {commandchars=\\\{\},fontsize=\footnotesize,breaklines=true,breakindent=8pt}

\providecommand{\tightlist}{%
  \setlength{\itemsep}{0pt}\setlength{\parskip}{0pt}}

\DeclareRobustCommand{\ttsplit}[1]{\texttt{\seqsplit{#1}}}

\usepackage{dblfloatfix}
\usepackage{placeins}
\usepackage{float}
\usepackage{multicol}
\newcommand{\openappendixcolumns}{\onecolumn\begin{multicols}{2}}
\newcommand{\closeappendixcolumns}{\end{multicols}\twocolumn}
\usepackage{needspace}
\newsavebox{\appendixwidebox}
\newdimen\appendixwideheight
\newcommand{\appendixwidetable}{\par\addvspace{0.6\baselineskip}\centering}
\newcommand{\appendixwidegap}{\par\addvspace{1.4\baselineskip}}
\newenvironment{appendixwide}
  {\begin{lrbox}{\appendixwidebox}\begin{minipage}{\textwidth}}
  {\end{minipage}\end{lrbox}%
   \par\addvspace{\baselineskip}%
   \appendixwideheight=\dimexpr\ht\appendixwidebox+\dp\appendixwidebox
                              +2\baselineskip\relax
   \ifdim\appendixwideheight<\textheight
     \Needspace*{\appendixwideheight}%
   \fi
   \noindent\usebox{\appendixwidebox}\par\addvspace{\baselineskip}}

\usepackage{xurl}

\hypersetup{
  pdftitle={Protocol Compression Changes Which Party Pays: Bilateral Cost
            in Cross-Organization LLM Agent Communication},
  pdfauthor={Janghoon Lee},
  pdfsubject={Agent communication; protocol negotiation; bilateral cost;
              tokenizer divergence; caching},
  pdfkeywords={LLM agents, protocol negotiation, communication cost,
               tokenizer divergence, cache pricing}}
\title{Protocol Compression Changes Which Party Pays: Bilateral Cost in
Cross-Organization LLM Agent Communication}
\author{Janghoon Lee\,\orcidlink{0009-0002-8108-5407} \\
        Redrob \\
        \texttt{janghoon@redrob.io} \\
        {\small ORCID 0009-0002-8108-5407}}

\begin{document}
\maketitle

\begin{abstract}
Agents that talk across organizations exchange long messages billed by the token. A shorter notation therefore looks like a saving that costs nothing but an agreement to use it.

Recent work reports the saving is conditional. Compressed notation can instead raise total tokens by 8\% to 11\% over a JSON baseline, when parsing failures force extra model calls. That is measured for one payer. Between two organizations neither side can install a decoder at the other end, and each pays under its own tokenizer, price, and cache state.

We measure both sides. A preregistered token-level study covered 198 content-matched item pairs across six vendors, for 2,376 native-usage cells. We then overlay an English baseline, runtime schema negotiation followed by compression, and injected-schema compression on a two-party procurement bargain with an exactly enumerated feasible set. The overlay covers 1,053 completed dialogues of a 1,215-cell grid across 3 model pairs, plus a 405-dialogue rerun of the negotiated condition.

Compression amplifies cross-vendor cost dispersion by a factor of 1.078, with a 95\% CI of {[}1.066, 1.091{]}, and two vendor pairs reverse which endpoint is cheaper. Runtime negotiation succeeds as a protocol and fails as a bargain. The parties agree a schema in 121 of 135 headline dialogues, none of them the schema we would have supplied. They settle the task in only 9 of those dialogues, and they reach impasse in 106 of them. The negotiated sessions average 10.8 turns against 17.6, and cost 52\% of the English total because sessions end sooner, not because the handshake is repaid. Break-even horizons run from 20 to 70 turns, the low end only under the conditional accounting, and all of them lie above every observed English session. On one cross-vendor pair both parties keep about half their cost. On the other the receiving party pays more at a high cache-hit rate.

\end{abstract}

\hypertarget{introduction}{%
\section{Introduction}\label{introduction}}

An agent proposes a shorter notation that reduces total tokens. Should the other agent accept? Existing compression results make this look like a technical choice. In a cross-organization channel it is an economic one, because each organization pays its own bill, uses its own tokenizer, and bears its own repair cost.

Aggregate savings do not answer the acceptance question. Let party A save more than it spends on negotiation while party B does not: the protocol is efficient under a pooled objective and unacceptable under bilateral rationality. The same wire string can also change relative cost across tokenizers, so compression can redistribute cost even when it reduces the sum. This paper studies the channels where that matters, the ones neither party fully controls, in which one party cannot install a deterministic decoder at the other endpoint, enforce a grammar on the other model, or assume the other endpoint's identity.

The object of study is a pair of net benefits, not one compression ratio:

\begin{equation*}
\begin{aligned}
\bigl( & \mathrm{benefit}_A(\mathrm{protocol}, N, h_A), \\
       & \mathrm{benefit}_B(\mathrm{protocol}, N, h_B) \bigr).
\end{aligned}
\end{equation*}

Adoption is mutually rational only where both components are positive, and a proposal outside that set should be rejected even if it lowers aggregate cost. Table \ref{tab:claims} states the four claims that follow, numbered (i) to (iv) and referred to by those numbers throughout.

\begin{table}[H]
\centering
\scriptsize
\setlength{\tabcolsep}{3pt}
\caption{Claim-status summary: each claim, the evidence this paper brings, its status, and where it is reported. This is the single place that states each status, and later sections point here rather than repeat the caveat.}\label{tab:claims}
\begin{tabularx}{\columnwidth}{@{}>{\hsize=1.300\hsize\RaggedRight\arraybackslash}X>{\hsize=0.900\hsize\RaggedRight\arraybackslash}X>{\hsize=0.900\hsize\RaggedRight\arraybackslash}X>{\hsize=0.900\hsize\RaggedRight\arraybackslash}X@{}}
\toprule
\textbf{Claim} & \textbf{Evidence} & \textbf{Status} & \textbf{Where} \\
\midrule
(i) The same wire string redistributes cost between endpoints, so compression has a per-party sign, not only a magnitude & P0b token-level dispersion & measured & §6 \\
(ii) That redistribution can reverse which party is cheaper when the representation changes & P0b, two vendor-pair reversals & measured & §6 \\
(iii) Direction-specific representations can expand the set of protocols both parties accept & C2 not run & unmeasured & Limitations \\
(iv) Negotiating into a compressed protocol at runtime can repay its own cost & C0/C1/C3 billed overlay & partial, at billed cost & §7, §8 \\
\bottomrule
\end{tabularx}
\end{table}

A preregistered token-level study, P0b, supplies the evidence for (i) and (ii). It replaced a first registered gate that failed, and we neither moved that threshold nor selected a passing subset (Section 6.1). The C0/C1/C3 overlay then addresses (iv) in billed conversation cost. Section 2 maps the prior results this work proceeds from, Section 3 fixes the channel boundary, Section 4 defines bilateral cost, Section 5 gives the design, and Sections 6 to 8 report P0b, the overlay, and the break-even region.

\hypertarget{related-work}{%
\section{Related Work}\label{related-work}}

Prior results supply the premises of this study rather than its targets. Runtime protocol negotiation already works. Agora \citep{agora} reports roughly 5-fold cost reduction, mainly by replacing repeated LLM work with deterministic routines, on single-round exchanges. Its switching values of 3, 5, and 10 are demonstration constants, which we reuse as fixed thresholds rather than claimed optima. Compressed notation can also backfire. Notation Matters \citep{notation-matters} turns per-call gains into total-token increases of 8\% to 11\% over a JSON baseline once parsing failures force extra inference iterations.

Caching, drift, and tokenization set the other three premises. Prompt caching cuts agent cost by 41\% to 80\% in synchronous workloads \citep{dont-break-the-cache}, and compression can break the prefix stability those savings rest on \citep{cache-aware-compression}. Multi-turn compliance degrades monotonically, at model-level violation rates from 8\% to 99\% \citep{multi-if} \citep{driftbench} \citep{sequor}, which we convert into party-specific expected repair cost rather than claim as a contribution. TensorZero \citep{tensorzero} measures Claude Opus at 5.3 times GPT on a tool workload against a 2-fold listed-price difference, with content-dependent rank changes for Gemini.

Each of those results is measured for one payer. Agora's exchanges are single round, Notation Matters fixes the format so negotiation costs nothing, the cache workloads are synchronous, and the drift benchmarks report violation rates rather than bills. Our change is to give the two endpoints separate tokenizers, prices, cache states, and adoption decisions. That separates five questions prior systems leave pooled: each party's cost, whether the mutual-benefit set can be empty, whether representation should differ by direction, what makes rejection rational on cost grounds, and how an asynchronous channel behaves when the gap between turns exceeds the cache lifetime. Separating them is the contribution of the framing, not answering all five; Table \ref{tab:claims} records which are answered here. Relay fidelity, fixed-protocol evaluation, protocol complexity, communication pricing, negotiation environments, post-processing compression, and acceptance primitives sit on axes this design does not test, and Appendix \ref{app:related} takes each in turn.

\hypertarget{scope-uncontrolled-channels}{%
\section{Scope: Uncontrolled Channels}\label{scope-uncontrolled-channels}}

We define an \textbf{uncontrolled channel} as communication for which at least one of the following is true:

\begin{enumerate}
\def\labelenumi{\arabic{enumi}.}
\tightlist
\item
  A party cannot deploy code or a deterministic routine in the other party's execution environment.
\item
  A party cannot enforce constrained decoding or structured output at the other endpoint.
\item
  A party does not know the other endpoint's model identity and settings in advance.
\end{enumerate}

Cross-organization email is the clearest example; a third-party agent behind a public API and an agent reached through a marketplace intermediary also fit. In each case a sender can propose a notation but cannot guarantee how the receiver decodes it. The scope excludes same-organization multi-agent systems, same-vendor pipelines, and settings where both parties can deploy a schema before communication. Those systems coordinate on code, grammar, tokenizer, and cache policy, so their optimization problem is real but is not bilateral bargaining under incomplete information. The exclusion is deliberate. Repair, Not Improvement \citep{repair-not-improvement} shows that a grammar can structurally remove format violations, and a reviewer who can deploy the same constrained decoder at both endpoints has moved outside the claim boundary.

The unit of analysis is one directed message as paid by its receiver, so a symmetric wire protocol need not have symmetric costs and we report A and B separately before any aggregate. We do not claim that every cross-organization exchange should negotiate a protocol. Negotiation has a cost and can fail, and the question is whether observable conditions justify a proposal, an acceptance, a rejection, or a counterproposal.

\hypertarget{bilateral-cost-model}{%
\section{Bilateral Cost Model}\label{bilateral-cost-model}}

\hypertarget{input-cost-with-caching}{%
\subsection{Input cost with caching}\label{input-cost-with-caching}}

For party \texttt{i} at turn \texttt{t}, let \(\mathrm{prefix}_{i,t}\) be the cumulative history already present before the new message and let \(\mathrm{new}_{i,t}\) be the newly added tokens under party \texttt{i}'s tokenizer. Let \(h_i\) be the probability that the prefix is served from cache. Input cost is

\begin{equation*}
\begin{aligned}
C_{\mathrm{in},i}(t, h_i) ={}& \mathrm{price}_{\mathrm{in},i} \\
&\cdot \bigl((1 - h_i)\,\mathrm{prefix}_{i,t} \\
&\quad {}+ \mathrm{new}_{i,t}\bigr) \\
&+ \mathrm{price}_{\mathrm{cached},i} \cdot h_i \\
&\quad {}\cdot \mathrm{prefix}_{i,t}.
\end{aligned}
\end{equation*}

The model keeps every subscript. A and B can have different input prices, cached-input prices, tokenizers, cache-hit probabilities, and prefix lengths for the same wire history, and that separation is the decisive change from a single-payer cost model. A single pooled hit rate, or one cached-read coefficient applied to every vendor, would conceal transfers between the parties.

We sweep \(h_i\) over \texttt{\{0,\ 0.5,\ 0.75,\ 0.9,\ 0.99\}} only for vendors whose official cached-input price is documented. A vendor with \texttt{cached\_read\_per\_1m\ =\ null} is reported at \(h_i = 0\), is omitted from the sweep, and raises if \(h_i > 0\) is requested: unknown is not replaced by zero or by a shared fraction of \texttt{price\_in}. Prices and retrieval URLs are snapshotted (Appendix \ref{app:artifacts}), verified 2026-08-24.

\hypertarget{negotiation-compliance-and-repair}{%
\subsection{Negotiation, compliance, and repair}\label{negotiation-compliance-and-repair}}

Let \(C_{\mathrm{neg},i}\) be party \texttt{i}'s cost for the handshake and schema negotiation, \(p_{\mathrm{viol},i}\) its probability of a protocol violation, \(C_{\mathrm{repair},i}\) the expected cost of repairing one, and \(\delta_{\mathrm{turn},i}\) the English cost minus the compressed cost per turn. The linear break-even length is

\begin{equation*}
N^{*}_{\mathrm{lin},i} = \frac{C_{\mathrm{neg},i}}{\delta_{\mathrm{turn},i} - p_{\mathrm{viol},i} \cdot C_{\mathrm{repair},i}}.
\end{equation*}

No finite positive break-even point exists when the denominator is nonpositive, and that fact is party-specific: a protocol can have a finite break-even point for A and none for B.

The context-aware version drops the constant-turn-saving assumption and sums the per-turn difference over the horizon instead, so it captures growing histories, prefix caching, and format-dependent tokenization (Appendix \ref{app:costdetail}). Mutual adoption at horizon \texttt{N} requires both parties' cumulative differences to be negative, and the mutual-benefit set is empty when no tested protocol and horizon satisfy both inequalities.

Negotiation failure probability \texttt{q} adds expected failed-handshake cost before adoption. The model admits an analytic sweep over \texttt{q}, which this overlay does not report because the refusal, silence, and misunderstanding counts it needs were never logged (Appendix \ref{app:negofail}). Two further parts of the model are in Appendix \ref{app:costdetail}: why an expired cache is not a substitute for compression, which adds a TTL-exceedance term to the hit rate, and the rule that tokens, listed cost, cache-adjusted cost, negotiation cost, and repair cost are reported separately rather than as one total.

\hypertarget{asymmetric-protocols}{%
\subsection{Asymmetric protocols}\label{asymmetric-protocols}}

A protocol is a pair of representations, one for each direction, and a symmetric policy is the special case that forces them to be equal. That is not an implementation detail. Receiver-side tokenization evaluates A-to-B traffic in B's environment and B-to-A traffic in A's, so a symmetric representation is optimal only if the same choice minimizes the objective in both directions subject to outcome equivalence.

\hypertarget{experimental-design}{%
\section{Experimental Design}\label{experimental-design}}

\hypertarget{task-and-endpoints}{%
\subsection{Task and endpoints}\label{task-and-endpoints}}

The overlay runs one negotiation task under every communication condition, so a cost difference is attributable to the condition rather than to a different task. The design targeted Talk is Cheap \citep{talk-is-cheap} with TERMS-Bench \citep{terms-bench} as the alternative, and neither had a confirmed Apache 2.0 or MIT source tree (Appendix \ref{app:suites}), so the overlay runs on \texttt{negoenv}, a five-issue procurement environment whose 1,440 allocations are enumerated exactly. It is a license fallback, not a contribution. Difficulty grades (\texttt{wide}, \texttt{narrow}, \texttt{empty}) keep the same preference structure and vary the reservation cutoff so that \ttsplit{\textbar{}F\textbar{}/1440} lands in a target band.

Three pairs run the overlay: \texttt{mid} (OpenAI nano with Gemini Flash), \texttt{haiku-qwen} (Claude Haiku with Qwen Flash), and \texttt{nano-nano} (self-play). Headline tables use \texttt{mid} at cache-hit rate \texttt{h\ =\ 0}; haiku-qwen is a cross-vendor sensitivity pair and nano-nano a same-model upper-bound reference. Endpoints enter only after a C3 self-play compliance screen at an arbitrary, unregistered 0.5 cutoff (Appendix \ref{app:screening}), and Mistral is excluded for rate limiting and a 0.111 compressed-wire compliance in pilot v4. Every completion runs with reasoning disabled (\texttt{reasoning.effort=none}). The grid is 1215 dialogues (15 instances × 3 seeds × 3 difficulties × 3 pairs × 3 conditions, so one condition is 405), flattened and shuffled under seed 20260824 so that an abort cannot truncate whole pairs and leave a systematically biased remainder. Prompt construction is in Appendix \ref{app:costdetail}.

\hypertarget{conditions}{%
\subsection{Conditions}\label{conditions}}

The condition registry contains C0 through C5: an English baseline, handshake with symmetric compression, asymmetric compression, oracle schema injection, immediate compressed proposal, and handshake followed by English, in that order. The recoupment overlay reported here is \textbf{C0, C1, and C3 only}. C2 tests claim (iii) and C4 and C5 are the handshake net-cost contrasts; none of the three was run. Every condition uses the same task input and terminal success predicate, negotiated conditions add handshake tokens to both parties rather than treating the handshake as free, and a rejected proposal continues in English while retaining the cost of the rejected negotiation.

The first C1 overlay in \ttsplit{data/p1/dialogues.jsonl} did \textbf{not} negotiate a schema: the runner injected \texttt{CANONICAL\_SCHEMA} at the same site as C3 and switched to compression on a turn counter, with no handshake challenge and no propose-accept-reject loop. That file is retained as the unimplemented contrast rather than relabeled. C1 was re-run on 2026-08-25 (405 dialogues, shuffle seed 20260825) with harness-generated handshake verification, model-proposed schema text, harness-judged \texttt{SCHEMA\ ACCEPT} / \texttt{SCHEMA\ REJECT}, and compression only after \texttt{schema\_agreed}. C3 was not changed and the two files are not merged.

\hypertarget{outcomes}{%
\subsection{Outcomes}\label{outcomes}}

The verifier returns \texttt{outcome\_class} in \texttt{\{correct,\ search\_failure,\ error\_empty,\ error\_below\_reservation\}}. Outcome equivalence is the two \texttt{error\_*} classes only, which ask whether compression produced a wrong conclusion; \texttt{search\_failure} is exploration under a turn budget and is compared as a separate question. Conditions are not ranked by \texttt{correct} rate alone. P1 has no registered gate, so no P1 number in this paper is a preregistered threshold.

The per-dialogue record keeps seven cost fields for each party separately, measured in that party's own environment, plus the shared outcome. Compliance and repair cost require a compressed turn measured against a known grammar, so they are defined for C3 and empty for C0, which has no compressed turn, and for implemented C1, whose grammar is negotiated per dialogue (Limitations); those cells are null rather than zero. Cost and outcome equivalence are defined for every condition. The record layout, the sweeps over cache-hit rate and TTL exceedance, the switching-policy signal set, the analysis-integrity rules, and every departure from the two preregistrations are in Appendix \ref{app:costdetail}.

\hypertarget{p0-tokenizer-divergence}{%
\section{P0: Tokenizer Divergence}\label{p0-tokenizer-divergence}}

\hypertarget{the-preregistered-gate-failed}{%
\subsection{The preregistered gate failed}\label{the-preregistered-gate-failed}}

The P0 gate is \textbf{FAIL}. The preregistered compressed-dispersion threshold was at least 1.300 and the observed value was 1.280. We did not revise the threshold and we did not narrow the confirmatory set to a passing subset. The second gate component passed, with relative token rankings reversing for 4 vendor pairs, but the registration required both.

The failure changed the design more than a narrow pass would have. The registered statistic was maximum divided by minimum token count, a magnitude, while the claim it was meant to screen is an existence claim about sign: that a configuration exists in which no protocol benefits both parties. Large dispersion does not imply that claim and small dispersion does not refute it, because two parties can have nearly equal magnitudes with opposite signs. The correct response is a new registration rather than a reinterpretation, and Appendix \ref{app:p0gate} carries the counting method, the rest of the argument, and the post-hoc observations that followed.

\hypertarget{p0b-re-operationalization-and-preregistration}{%
\subsection{P0b: re-operationalization and preregistration}\label{p0b-re-operationalization-and-preregistration}}

P0b is registered separately, reuses no P0 item or count, and splits the two jobs the failed gate confused. The amplification ratio \(A = D_{\mathrm{comp}} / D_{\mathrm{eng}}\), where each \(D\) is the same-list maximum-to-minimum native prompt-token total, carries the magnitude. The directional screen \texttt{K}, the number of unordered vendor pairs whose English token-order sign disagrees with their compressed sign, carries the sign. \texttt{PASS\ =\ G1\ AND\ G3}: the 95\% interval lower bound for \texttt{A} must exceed 1.000, and at least one pair must conflict in at least 95\% of 10,000 paired bootstrap replicates. Both are hypothesis boundaries rather than constants taken from P0. The corpus is 198 new content-matched pairs, 33 in each of six notation families, hash-split into selection and confirmation halves; family-level numbers here use the confirmation half.

\hypertarget{p0b-results}{%
\subsection{P0b results}\label{p0b-results}}

The registered gate is \textbf{PASS}. Native \texttt{usage.prompt\_tokens} was recorded for all 2,376 planned cells (198 items × 6 vendors × 2 surfaces), so the missing rate is 0.0 against a registered 5\% data-quality stop. \texttt{A} is 1.078 with a 95\% interval of {[}1.066, 1.091{]}, clearing G1, and \texttt{K} is 2 in every bootstrap replicate, clearing G3. The two qualifying pairs are \texttt{alibaba/mistral} and \texttt{anthropic/google}, each at conflict rate 1.0, with the other 13 of 15 unordered pairs at 0.0, so the sign conflicts are not an artifact of a handful of resamples. The reporting-only spread \texttt{S} is 1.167 {[}1.146, 1.189{]}.

Vendor totals show why a single compression ratio is the wrong summary: Anthropic's compressed total exceeds its English total (1.056) while DeepSeek retains the smallest fraction (0.905), on the same 198 items. Pooled \texttt{A} and \texttt{K} are token counts and include no handshake, repair, listed price, cache, or horizon term. Appendix \ref{app:p0tables} carries the gate outcomes, the per-vendor totals, the pair conflict rates, and the family tables.

\hypertarget{unregistered-post-hoc-diagnostic-numeral-density-versus-symbols}{%
\subsection{Unregistered post-hoc diagnostic: numeral density versus symbols}\label{unregistered-post-hoc-diagnostic-numeral-density-versus-symbols}}

The P0b stimuli are numeral-dense and compression drops English filler, so pooled \texttt{A\ \textgreater{}\ 1} could come from digit packing rather than from exotic operators. \texttt{json\_baseline} is the control that keeps the numeric payload without exotic operators. Its confirmation-half \texttt{A} is 0.973 (95\% CI {[}0.967, 0.979{]}), an interval lying entirely below 1, and all five symbol families sit above its upper bound. The diagnostic is unregistered and is not a gate. It supports a split between the symbol families as a group and the numeric control, not a causal decomposition of \texttt{A} (Appendix \ref{app:p0tables}).

\hypertarget{results}{%
\section{Results}\label{results}}

The archived run \ttsplit{data/p1/dialogues.jsonl} stopped at the approved cap at \textbf{1053 / 1215} \texttt{status=ok}, and its C1 rows are the \textbf{unimplemented} injected-schema condition (Section 5.2), so headline C1 below is the 2026-08-25 renegotiation overlay \ttsplit{data/p1/dialogues\_c1\_nego.jsonl} (405/405 \texttt{status=ok}). Headline tables use pair \texttt{mid} (\texttt{openai/gpt-5.4-nano} × \ttsplit{google/gemini-2.5-flash}), with 10,000 bootstrap replicates and 95\% percentile intervals. \texttt{agreement} is both-party accept (\texttt{terminal==agreement}) only, impasse and turn-cap are separate exits, and protocol agreement is \texttt{schema\_agreed}, a different predicate.

\hypertarget{headline-listed-cost-length-and-exits-mid}{%
\subsection{\texorpdfstring{Headline listed cost, length, and exits (\texttt{mid})}{Headline listed cost, length, and exits (mid)}}\label{headline-listed-cost-length-and-exits-mid}}

\begin{table*}[tp]
\centering
\footnotesize
\caption{Headline listed cost, length, and exits for the \texttt{mid} pair: task agreement, impasse and turn-cap rates, party-sum listed cost at \texttt{h\ =\ 0} and \texttt{h\ =\ 0.9}, and mean turns. Point estimates; 95\% intervals and cost per turn are in Appendix \ref{app:overlay}.}\label{tab:headline}
\begin{tabular}{@{}lrrrrrr@{}}
\toprule
\textbf{Condition} & \textbf{agree} & \textbf{impasse} & \textbf{cap} & \textbf{h=0} & \textbf{h=0.9} & \textbf{turns} \\
\midrule
C0 English (n=118) & 0.00847 & 0.551 & 0.441 & 0.00735 & 0.00228 & 17.6 \\
C1 implemented (n=135) & 0.0667 & 0.785 & 0.148 & 0.00379 & 0.00112 & 10.8 \\
C3 oracle schema (n=112) & 0 & 0.821 & 0.179 & 0.00499 & 0.00152 & 12.3 \\
\bottomrule
\end{tabular}
\end{table*}

On listed total cost at \texttt{h\ =\ 0}, implemented C1 is \textbf{52\%} of C0 (0.00379 / 0.00735), and the same ordering holds at \texttt{h\ =\ 0.9} (49\%). That comparison spans two runs, so it is also computed matched on pair × difficulty × instance\_index × seed: matched C1 − C0 total is \textbf{−0.00355} {[}−0.00405, −0.00304{]} (n=118) against the unmatched difference of −0.00356, so the pairing does not move it. Matched C1 − C3 is −0.00122 {[}−0.00188, −0.000558{]} (n=112) and matched C3 − C0 is −0.00237 {[}−0.00298, −0.00175{]} (n=99), unchanged from the archived analysis. The unimplemented C1 row is kept as a labelled contrast rather than deleted: 0.0696 agreement, 0.930 impasse, 5.15 turns (n=115), from a condition that never ran a schema negotiation.

\hypertarget{protocol-gate-schema_agreed}{%
\subsection{\texorpdfstring{Protocol gate (\texttt{schema\_agreed})}{Protocol gate (schema\_agreed)}}\label{protocol-gate-schema_agreed}}

Implemented C1 agrees a schema in \textbf{121/135 = 0.896} headline dialogues and \textbf{285/405 = 0.704} overall, at pair rates \texttt{haiku-qwen} 0.985 and \texttt{nano-nano} 0.230. No agreed schema matches \texttt{CANONICAL\_SCHEMA} (0/285). Handshake verification is 85/405 = 0.210 overall and 61/135 = 0.452 on \texttt{mid}, and failure there does not block schema agreement: 206 of 320 failed handshakes still agree one. Task \texttt{agreement} on implemented headline C1 is 9/135 = 0.0667, the same order as unimplemented C1 (8/115 = 0.0696). Schema agreement is not task agreement.

The dialogues end in exits rather than silence. C3 agreement is 0/112 with 92 impasse and 20 turn\_cap, and implemented C1 is 106/135 impasse and 20/135 turn\_cap. Empty-F impasse is the correct conclusion, while wide or narrow impasse is \texttt{search\_failure}, at 0.778 under implemented C1 against 0.506 (C0) and 0.813 (C3). Appendix \ref{app:overlay} carries the rates by condition and pair, the outcome-class tallies, and the compliance figure.

\hypertarget{cheaper-for-whom}{%
\subsection{Cheaper for whom}\label{cheaper-for-whom}}

\begin{figure*}[tp]
\centering
\includegraphics[width=0.92\textwidth]{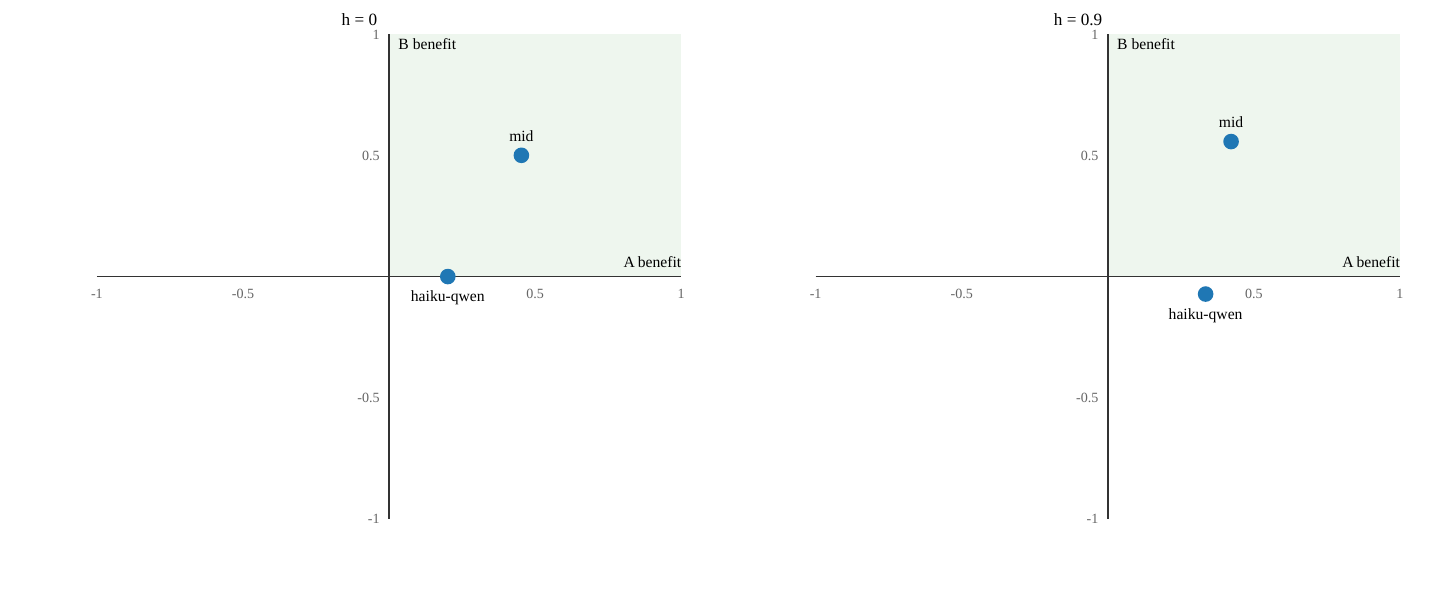}
\caption{Change in each party's billed cost from C0 to \textbf{implemented} C1, as a fraction of that party's C0 cost, so the axes are unitless and positive means the party pays less. One point per cross-vendor pair at \texttt{h\ =\ 0} and \texttt{h\ =\ 0.9}. The shaded quadrant is where both signs are positive. Self-play is excluded because it is not a bilateral pair. Drawn from \ttsplit{data/p1/dialogues\_c1\_nego.jsonl} against archived C0.}\label{fig:cheaper}
\end{figure*}

The claim is about sign, and the sign splits by party. On \texttt{mid} both parties pay less (A 0.453, B 0.501 at \texttt{h\ =\ 0}). On \texttt{haiku-qwen} A pays 0.201 less while B is unchanged at 0.0002, and at \texttt{h\ =\ 0.9} B pays 0.072 \textbf{more} against A's 0.335 less. Token-level result (ii) found the cheaper party reversing between surfaces, and the same divergence in sign survives into billed conversation cost. This does not establish an empty mutual-benefit set: that predicate compares protocols at a held outcome, and these two conditions hold neither length nor outcome fixed.

\hypertarget{summary-of-the-overlay}{%
\subsection{Summary of the overlay}\label{summary-of-the-overlay}}

Implemented C1 forms a schema in most headline dialogues, never adopts the canonical injection, and settles the task in 6.7\% of them. Its listed total is 52\% of C0 because the session is shorter, not because handshake plus schema-nego cost is recovered inside a C0-length bargain. On point means of total cost alone the analysis plan would read this as ``compression helps and negotiation is recouped'', which is the wrong reading (Appendix \ref{app:overlay}). The overlay records protocol success with task failure.

\hypertarget{interpretive-break-even-region}{%
\section{Interpretive Break-Even Region}\label{interpretive-break-even-region}}

The planned online policy comparison was not performed. The reduced overlay has no condition space for it, and agreement stays below 7\% in every headline condition, so the data identify no policy decision boundary. This section reports an interpretive break-even calculation on measured C0, C1, and C3 costs for the headline \texttt{mid} pair instead. It is not a policy evaluation.

\hypertarget{measured-inputs}{%
\subsection{Measured inputs}\label{measured-inputs}}

For party \texttt{i}, let \(c_{0,i}\) be the mean C0 cost per turn and let \(d_i\) be the compressed-to-English per-turn ratio, the mean C3 cost per turn over that same quantity under C0. The ratio does not depend on dialogue length. Let \(C_{\mathrm{neg},i} = \operatorname{mean}(C_{\mathrm{handshake},i} + C_{\mathrm{schema},i})\) be party \texttt{i}'s mean implemented-C1 cost on the runner-labeled \texttt{handshake} and \texttt{schema\_nego} turns, excluding compressed turns and post-failure English task turns. The interpretive horizon is

\begin{equation*}
N^{*}_i = \frac{C_{\mathrm{neg},i}}{c_{0,i}\,(1 - d_i)}.
\end{equation*}

Cache-hit rate follows the registered sweep \texttt{\{0,\ 0.5,\ 0.75,\ 0.9,\ 0.99\}}. Expected \(C_{\mathrm{neg}}\) averages all headline implemented C1 dialogues, conditional \(C_{\mathrm{neg}}\) averages \texttt{schema\_agreed} dialogues only, and Appendix \ref{app:overlay} records an earlier draft that set the numerator to C1 − C3 and so mixed negotiation cost with a length difference.

\hypertarget{party-specific-horizons}{%
\subsection{Party-specific horizons}\label{party-specific-horizons}}

Implemented headline C1 (\texttt{mid}, n=135) has mean handshake 2 turns and mean \texttt{schema\_nego} 2.35 turns. At \texttt{h\ =\ 0}, expected \(C_{\mathrm{neg}}\) is 0.000513 (A) and 0.000816 (B), the conditional values are 0.000475 and 0.000736, and \(d\) is 0.949 for A and 0.944 for B. Table \ref{tab:nstar} sweeps the result: \(N^*\) is positive and \textbf{larger than} the measured C0 mean of 17.6 turns at every tested \texttt{h}, for both parties, under both averages. Figure \ref{fig:breakeven} in Appendix \ref{app:overlay} plots it.

\begin{table}[H]
\centering
\scriptsize
\setlength{\tabcolsep}{3pt}
\caption{Interpretive break-even horizon \(N^*\) by cache-hit rate for headline implemented C1. Expected (exp.) averages all headline dialogues; conditional (cond.) averages \texttt{schema\_agreed} dialogues only. Every entry exceeds the measured C0 mean of 17.6 turns, so no row is recouped at the observed English length.}\label{tab:nstar}
\begin{tabularx}{\columnwidth}{@{}>{\hsize=0.900\hsize\RaggedLeft\arraybackslash}X>{\hsize=1.003\hsize\RaggedLeft\arraybackslash}X>{\hsize=1.003\hsize\RaggedLeft\arraybackslash}X>{\hsize=1.047\hsize\RaggedLeft\arraybackslash}X>{\hsize=1.047\hsize\RaggedLeft\arraybackslash}X@{}}
\toprule
\textbf{h} & \textbf{exp. \(N^*_A\)} & \textbf{exp. \(N^*_B\)} & \textbf{cond. \(N^*_A\)} & \textbf{cond. \(N^*_B\)} \\
\midrule
0 & 70.17 & 54.5 & 64.94 & 49.12 \\
0.5 & 64.97 & 46.31 & 60.56 & 41.07 \\
0.75 & 60.18 & 38.48 & 56.54 & 33.38 \\
0.9 & 55.77 & 30.98 & 52.83 & 26.01 \\
0.99 & 52.15 & 24.63 & 49.79 & 19.77 \\
\bottomrule
\end{tabularx}
\end{table}

Listed total C1 cost is still below C0 because implemented C1 is shorter and usually ends in impasse. That gap is not recoupment of handshake plus schema negotiation inside a C0-length English bargain. The unimplemented C1 corpus had \(C_{\mathrm{neg}} = 0\) because those phases were absent on the wire, which is a runner defect rather than evidence that bargaining was free.

\hypertarget{when-compression-can-pay}{%
\subsection{When compression can pay}\label{when-compression-can-pay}}

Compression can pay when the conversation lasts long enough to amortize a positive negotiation cost, the compressed representation reduces per-turn cost enough for both parties, and cache reads arrive before expiry. The first condition fails here: \(d < 1\) at every tested \texttt{h}, so a finite horizon exists, but it lies beyond the observed English session. The third is untested on this pair, because neither headline endpoint documents both a numeric TTL and a write premium (Appendix \ref{app:cache}). Agora's thresholds of 3, 5, and 10 are demonstration constants rather than optima, and a derived threshold moves outside that range when prices, tokenization, cache state, or negotiation cost change.

\hypertarget{discussion}{%
\section{Discussion}\label{discussion}}

Preserving the P0 failure is what exposed the distinction between cost magnitude and benefit sign, and bilateral adoption depends on the sign. P0b registers that distinction, as token-level evidence that a wire string can redistribute cost rather than as a conversation-cost mutual-benefit set.

High protocol compliance does not explain away the C3 exit pattern. The headline \texttt{mid} C3 cell has mean compliance 0.994 for A and 0.998 for B and no agreement among 112 dialogues, and 89 of the 109 dialogues with minimum party compliance at least 0.9 end in impasse. That rules out low compliance as a necessary condition for the C3 fold without identifying its cause. The pooled association between low compliance and impasse is weaker than it looks, because the two rates come from different pairs (Appendix \ref{app:overlay}).

Model composition changes termination behavior. The headline wide and narrow impasse gap is 30.7 percentage points and falls to 7.9 after pooling three pairs, and implemented C1 \texttt{schema\_agreed} depends on pair choice the same way (\texttt{nano-nano} 0.230 against cross-vendor rates above 0.89). That is consistent with result (ii): party composition can change a bilateral conclusion, though three pairs do not establish a model taxonomy. Running C2 would test claim (iii), and C4 and C5 would identify handshake net cost.

\hypertarget{conclusion}{%
\section{Conclusion}\label{conclusion}}

Protocol adoption in a cross-organization channel is a bilateral decision, and this study measures both sides of it. The preregistered token-level study establishes that the same wire string redistributes cost between endpoints, with a registered amplification above 1 and two vendor pairs whose order reverses. That is the premise the overlay tests in conversation cost, within the uncontrolled channels of Section 3.

Runtime negotiation succeeds as a protocol and fails as a bargain. The implemented condition agrees a schema in most headline dialogues, never adopts the canonical injection, and still settles the task in fewer than 7\% of them. Its listed total is about half of the English baseline because the session ends sooner, mostly in impasse, not because the handshake and schema negotiation are earned back inside a baseline-length exchange, and every interpretive horizon exceeds the measured English session length.

Whether compression pays also depends on which pair is asked. On the headline pair both parties keep a substantial share of their cost; on the other cross-vendor pair the gain is one-sided, and at a high cache-hit rate the receiving party is worse off. A single aggregate saving would have hidden that. The overlay identifies how cost is redistributed, not what a successful bargain costs each party.

\hypertarget{sec:limitations}{%
\section*{Limitations}\label{sec:limitations}}
\addcontentsline{toc}{section}{Limitations}

\textbf{Retried turns are kept in every table.} A retried completion can differ from the first attempt, so records carry \texttt{retried}, \texttt{retry\_count}, and \texttt{retry\_reason}. The rate is low in both overlays: 130 of 13,363 turns (0.97\%) in the archived run and 53 of 5,340 (0.99\%) in the implemented C1 run, across 100 of 1,053 and 29 of 405 dialogues, with at most two retries on any turn. The reasons are rate limiting (115 and 49) and transient transport errors (15 and 4), not model behaviour. We keep those turns because the cost tables are native usage actually billed, and a retry is part of that bill. Dropping them would remove paid tokens from a cost measurement.

\textbf{The first P1 main run was aborted and is not used for analysis.} On 2026-08-24 the run stopped at the approved cap (\texttt{13.9035} USD). In-flight dialogues finished at \texttt{14.179} USD native \texttt{usage.cost} (overrun \texttt{0.276} USD, recorded as spent, not adjusted). Completions were pair-sequential: mid 810/810, cheap 603/810, haiku-deepseek 383/810, and three pairs unstarted. That remainder is systematically biased by vendor pair, not missing at random. Those three counts are \texttt{status=ok} records and sum to 1796; the archived file holds 2003 rows because it also keeps the 207 cells that were written as \texttt{missing} rather than dropped. It is in the aborted-run archive (Appendix \ref{app:artifacts}) and is not merged with any later run.

\textbf{OpenRouter's listed OpenAI cache-read multipliers (0.25× / 0.50×) are not the sweep prices.} The \texttt{h} sweep keeps official vendor list prices (OpenAI nano cached input \$0.02 / 1M = 0.10× listed prompt). Mixing the two sources would make the cost basis unidentifiable.

\textbf{The P0 gate failed.} Compressed dispersion did not reach the preregistered threshold. The benefit-spread and amplification statistics were selected after observing P0 and are descriptive. They require a new registration and held-out data.

\textbf{The vendor list is finite.} P0 uses 6 vendor families. Maximum-to-minimum dispersion depends on which endpoints enter that list. Adding or removing a vendor can change both extremes without changing any existing observation.

\textbf{Rank reversal is not conversation cost.} P0 reversals concern native input token counts. They do not include listed price, cache writes, cache reads, output tokens, handshake cost, violations, repairs, or conversation horizon. They are directional evidence, not proof of an empty mutual-benefit set.

\textbf{Negotiation failure categories were not logged.} Appendix \ref{app:negofail} fills C1 phase counts, schema-confirmation rates, phase costs, and terminal reasons from \ttsplit{data/p1/dialogues.jsonl}. Refusal, silence, and misunderstanding remain unmeasured. The planned 20-instance taxonomy file was not collected, and no analytic-\texttt{q} figure is substituted.

\textbf{Same-model pairs are not the headline.} They provide an upper-bound reference for coordination under matched environments. They do not represent the cross-organization uncertainty that motivates the paper.

\textbf{The cache mechanism is a parameter sweep.} An expired write has no future read value by construction. TTL exceedance \texttt{e} is swept like \texttt{q} rather than treated as a missing prevalence measurement. The cache-horizon term is a cost-model implication over that sweep, not a claim about how often real organizations miss the TTL. TTL and write premiums are vendor-specific. Anthropic documents 5 min at 1.25× and 1 h at 2×. OpenAI nano, Google implicit, Alibaba implicit, Mistral, and DeepSeek V3.1 do not jointly document both fields. The write-without-read premium table omits them rather than assign another vendor's ratios.

\textbf{DeepSeek V3.1 cached-input USD is unverified.} The official DeepSeek pricing page current on 2026-08-24 lists V4 cache hit/miss prices. Those rates are not applied to \ttsplit{deepseek/deepseek-chat-v3.1}. That endpoint is therefore held at \texttt{h\ =\ 0} wherever it appears, including in tables whose other rows sweep \texttt{h\ \textgreater{}\ 0}, and Section 4.1 makes the implementation raise rather than substitute a cached price. It is not in the headline pair, so it does not enter Figure \ref{fig:breakeven}.

\textbf{Complete-case bootstrap can be biased if missingness correlates with vendor.} The paired item bootstrap drops any item that lacks a native usage cell for every vendor and surface. If missingness is vendor-correlated, the complete-case estimand need not equal the full-data estimand. The registered 5\% missing-rate stop limits how large that bias can grow before the study is declared a data-quality failure rather than a G1/G3 result.

\textbf{Protocol families are heterogeneous.} The Round 2 family values show that symbol-dense formats do not form one treatment. Only \texttt{unicode\_ops} has selection in Round 1 and confirmation in Round 2. The other family-level findings are exploratory.

\textbf{The P0b post-hoc digit-versus-symbol contrast is unregistered.} It is not confirmatory. \texttt{json\_baseline} supplies a lower bound on the numeral contribution, not a complete decomposition of \texttt{A}. Adjacent residual ranks among symbol families are not established: \texttt{pipe\_kv} and \texttt{set\_builder} have overlapping confirmation CIs. What is supported is the split between the five symbol families as a group and the numeric JSON control.

\textbf{P0b stimuli are synthetic and single-domain.} Every item is a records/limit/threshold template. Generalization to naturally occurring agent traffic is untested.

\textbf{K = 2 is a token-count statistic.} It does not establish an empty mutual-benefit set in actual conversation cost. That claim still requires P2.

\textbf{The negotiation substrate is in-house.} \texttt{negoenv} was assembled because no Apache 2.0 or MIT evaluation loop was available (Appendix \ref{app:suites}). Scores are not comparable to TERMS-Bench or Talk is Cheap leaderboards.

\textbf{The issue space is a single procurement domain.} Five issues (seats, unit price, term, support tier, start offset) do not represent other bargaining settings.

\textbf{Difficulty grades change the reservation, not the preference structure.} \texttt{wide}, \texttt{narrow}, and \texttt{empty} share statistically the same weights, value slopes, and conflict intensity. Reservations are fit so that \ttsplit{\textbar{}F\textbar{}/1440} hits a band. The grades scale the agreement set; they do not change structural bargaining difficulty.

\textbf{Completions are non-reasoning.} Hidden chain-of-thought is off (\texttt{reasoning.effort=none}) for every P1 call. Token costs therefore include only the native usage of that mode. Results do not transfer to reasoning-on settings.

\textbf{Terminal reminders are on every turn.} English and compressed messages end with a one-line restatement of offer/accept/impasse tokens and remaining turns. Outcomes are for that reminder regime. We did not test whether the same terminals appear without the suffix.

\textbf{\texttt{search\_failure} mixes search skill and turn budget.} We do not separate those. A nonempty F with an impasse or a timeout is scored as search failure, not as a wrong conclusion.

\textbf{C2, C4, and C5 were omitted from the recoupment overlay.} Claim (iii) is asymmetric, direction-specific compression, which remains unmeasured because C2 was not run. Party-specific break-even under cache and horizon belongs to claim (iv), which Section 8 addresses at billed cost only. Handshake net cost in English (C5 − C0) and under skipped handshake (C4 − C3) also remain unmeasured. The archived 6×6 remainder (Appendix \ref{app:artifacts}), 192 rows spread across all six pairs and all six conditions, is not used for those claims. It is a separate archive from the aborted run described above.

\textbf{The planned switching-policy comparison was not performed.} The reduced C0/C1/C3 overlay lacks the required condition space. It cannot compare an online policy with fixed thresholds, always accept, never accept, and an oracle. Agreement below 7\% in every headline condition also leaves no empirical decision boundary. Section 8 reports measured cost arithmetic instead of policy performance.

\textbf{Table \ref{tab:planmap} uses total cost alone.} It selects the registered C3 \textless{} C0 and C1 \textless{} C0 cell. That rule does not require outcome equivalence. The selected sentence therefore cannot serve as the behavioral conclusion when acceptances disappear and impasses rise.

\textbf{The overlay stopped at the spend cap.} Completions are 1053/1215. The job list was shuffled (seed 20260824), so coverage is roughly balanced (pairs 345/350/358; conditions 360/344/349), not a pair-truncated slice. The 162 unfinished cells are not missing at random in calendar time; they are the tail of a shuffle. We do not raise the cap. The balanced 1053-row sample is the final overlay record.

\textbf{The first C1 overlay did not negotiate.} \ttsplit{data/p1/dialogues.jsonl} C1 injected the canonical schema and compressed after two English turns. That is recorded, not relabeled. Implemented C1 is \ttsplit{data/p1/dialogues\_c1\_nego.jsonl} (405 dialogues, 2026-08-25). The files are not merged.

\textbf{The bargaining task is hard for these endpoints.} Headline \texttt{mid} task \texttt{agreement} is 0.85\% (C0), 6.67\% (implemented C1), 0\% (C3), and 6.96\% (unimplemented C1 contrast). Implemented C1 \texttt{schema\_agreed} is 89.6\% on that pair. Protocol agreement is not task agreement.

\textbf{A negotiated protocol needs a per-dialogue verifier, and this study did not build one.} Compliance here is a wire tested against one fixed grammar, the canonical field set. That predicate does not apply to implemented C1, where every one of the 285 agreed schemas differs from the canonical text, so the correct object to check is each dialogue's own \texttt{schema\_text} and the correct instrument is a verifier generated from it. No such generator exists in this harness, and the runner (Appendix \ref{app:artifacts}) skips the check for C1 rather than applying the wrong grammar. Every C1 turn is therefore recorded as compliant by construction, which is an absent measurement and not a perfect score. C1 is left out of Figure \ref{fig:compliance}, and its repair-cost cells are null rather than zero. This is a missing instrument on the same footing as the omitted conditions, not an implementation slip: measuring compliance under runtime negotiation requires building schema-to-verifier compilation, and whether a self-authored schema is followed more faithfully than an injected one stays open until it exists.

\textbf{A retried turn is one turn and one billed completion.} Retries fire on a rejected or failed request, which returns no usage record, so the recorded cost and token counts come from the attempt that succeeded. Turn counts and cost per turn are therefore unaffected by the retry rate. A provider that charges for a rejected attempt would be invisible to this accounting.

\textbf{Retry pressure is not spread evenly across the parties.} Of the 130 retried turns in the archived run, 119 fall on party B and 116 of those on the Alibaba endpoint; the implemented C1 run repeats the pattern with 51 of 53 on party B and 50 on the same endpoint. Nearly all are 429s, and they concentrate in the \texttt{haiku-qwen} pair. Wall-clock pacing on that pair is therefore a property of one vendor's rate limit rather than of the protocol, which is the same reason Mistral was excluded from the overlay.

\textbf{Same-model self-play is not an upper bound on protocol agreement.} On implemented C1, \texttt{nano-nano} \texttt{schema\_agreed} is 0.230 against 0.896 (\texttt{mid}) and 0.985 (\texttt{haiku-qwen}). Headline protocol rates therefore depend on which pair is selected. Self-play remains an upper-bound reference for matched environments only, not for schema agreement. We did not measure why the same-model pair agrees less often.

\hypertarget{sec:ethics}{%
\section*{Ethics Statement}\label{sec:ethics}}
\addcontentsline{toc}{section}{Ethics Statement}

This study has no human subjects, no annotation task, and no personal data. Every item is generated from templates by \texttt{negoenv} and the P0b stimulus script. No model was trained or fine-tuned, so the compute is the inference already billed below.

All measurement was paid inference on commercial endpoints reached through OpenRouter. The recorded spend is 0.138 USD for the P0b token counts, 14.18 USD for the aborted first P1 run, 9.36 USD for the archived overlay, and 3.25 USD for the C1 rerun. Each run carried an approved cap that stopped it, and the archived overlay is reported at the 1,053 dialogues that cap allowed rather than extended to fill the grid.

Vendor prices in this paper are public list prices snapshotted on 2026-08-24 with their retrieval URLs. They move, and a reader who reuses a number here without re-fetching it is reading a dated quote. Where a price was not documented we left the cell empty rather than transferring another vendor's ratio, which is why several endpoints are held at \texttt{h\ =\ 0}.

The cost orderings reported here are not quality judgments. A vendor that tokenizes a compressed surface into more tokens is not a worse model, and nothing in this design measures answer quality, safety, or fitness for any task. The token-order reversals are a property of a tokenizer and a wire format together, on one synthetic single-domain corpus, and they should not be cited as a vendor ranking.

Two candidate negotiation suites were not vendored because their license or source tree could not be confirmed as Apache 2.0 or MIT (Appendix \ref{app:suites}). We used our own substrate instead of copying material we had no clear right to redistribute. The inquiry we drafted to one set of authors was never sent, and this paper makes no claim about their intent.

\clearpage

% ACL order: the numbered body and the unnumbered Limitations and Ethics
% sections, then the references, then the appendix. 001 and 002 set the same
% order. acl.sty already selects acl_natbib, and a second \bibliographystyle
% is an error rather than an override.
\bibliography{refs}

\begin{thebibliography}{18}
\providecommand{\natexlab}[1]{#1}

\bibitem[{Canaverde et~al.(2026)Canaverde, Alves, Pombal, Attanasio, and
  Martins}]{sequor}
Beatriz Canaverde, Duarte~M. Alves, José Pombal, Giuseppe Attanasio, and
  André F.~T. Martins. 2026.
\newblock \href {https://arxiv.org/abs/2605.06353} {Sequor: A multi-turn
  benchmark for realistic constraint following}.
\newblock \emph{Preprint}, arXiv:2605.06353.

\bibitem[{Dobrovolskyi(2026)}]{mcp-vs-a2a}
Ivan Dobrovolskyi. 2026.
\newblock \href {https://arxiv.org/abs/2603.22823} {Empirical comparison of
  agent communication protocols for task orchestration}.
\newblock \emph{Preprint}, arXiv:2603.22823.

\bibitem[{Du et~al.(2025)Du, Su, Li, Ding, Yang, Han, Tang, Zhu, and
  You}]{protocolbench}
Hongyi Du, Jiaqi Su, Jisen Li, Lijie Ding, Yingxuan Yang, Peixuan Han, Xiangru
  Tang, Kunlun Zhu, and Jiaxuan You. 2025.
\newblock \href {https://arxiv.org/abs/2510.17149} {Protocolbench: Which llm
  multiagent protocol to choose?}
\newblock \emph{Preprint}, arXiv:2510.17149.

\bibitem[{Eslami(2026)}]{communication-pricing}
Mojtaba Eslami. 2026.
\newblock \href {https://arxiv.org/abs/2608.07532} {Dynamic coalition formation
  and communication pricing in skill-based agentic ai systems}.
\newblock \emph{Preprint}, arXiv:2608.07532.

\bibitem[{He et~al.(2024)He, Jin, Wang, Bi, Mandyam, Zhang, Zhu, Li, Xu, Lv,
  Bhosale, Zhu, Sankararaman, Helenowski, Kambadur, Tayade, Ma, Fang, and
  Wang}]{multi-if}
Yun He, Di~Jin, Chaoqi Wang, Chloe Bi, Karishma Mandyam, Hejia Zhang, Chen Zhu,
  Ning Li, Tengyu Xu, Hongjiang Lv, Shruti Bhosale, Chenguang Zhu,
  Karthik~Abinav Sankararaman, Eryk Helenowski, Melanie Kambadur, Aditya
  Tayade, Hao Ma, Han Fang, and Sinong Wang. 2024.
\newblock \href {https://arxiv.org/abs/2410.15553} {Multi-if: Benchmarking llms
  on multi-turn and multilingual instructions following}.
\newblock \emph{Preprint}, arXiv:2410.15553.

\bibitem[{{IBM Research}(2025)}]{acp}
{IBM Research}. 2025.
\newblock Agent communication protocol {(ACP)}.
\newblock \url{https://github.com/i-am-bee/acp}.
\newblock Open specification, Apache-2.0, now governed by the Linux Foundation.
  Repository archived after the merge into A2A.

\bibitem[{Kruthof(2026)}]{driftbench}
Garvin Kruthof. 2026.
\newblock \href {https://arxiv.org/abs/2604.28031} {Models recall what they
  violate: Constraint adherence in multi-turn llm ideation}.
\newblock \emph{Preprint}, arXiv:2604.28031.

\bibitem[{Kutschka and Geiger(2026)}]{notation-matters}
Lorenz Kutschka and Bernhard Geiger. 2026.
\newblock \href {https://arxiv.org/abs/2605.29676} {Notation matters: A
  benchmark study of token-optimized formats in agentic ai systems}.
\newblock \emph{Preprint}, arXiv:2605.29676.

\bibitem[{Lee(2026)}]{repair-not-improvement}
Janghoon Lee. 2026.
\newblock \href {https://arxiv.org/abs/2608.13959} {Repair, not improvement:
  Decomposing constrained decoding in tool-call abstention}.
\newblock \emph{Preprint}, arXiv:2608.13959.

\bibitem[{{LF AI \& Data}(2025)}]{acp-a2a-merge}
{LF AI \& Data}. 2025.
\newblock {ACP} joins forces with {A2A} under the {Linux Foundation}'s {LF AI
  \& Data}.
\newblock
  \url{https://lfaidata.foundation/communityblog/2025/08/29/acp-joins-forces-with-a2a-under-the-linux-foundations-lf-ai-data/}.
\newblock Announcement of 2025-08-29.

\bibitem[{Lumer et~al.(2026)Lumer, Nizar, Jangiti, Frank, Gulati, Phadate, and
  Subbiah}]{dont-break-the-cache}
Elias Lumer, Faheem Nizar, Akshaya Jangiti, Kevin Frank, Anmol Gulati, Mandar
  Phadate, and Vamse~Kumar Subbiah. 2026.
\newblock \href {https://arxiv.org/abs/2601.06007} {Don't break the cache: An
  evaluation of prompt caching for long-horizon agentic tasks}.
\newblock \emph{Preprint}, arXiv:2601.06007.

\bibitem[{Marro et~al.(2024)Marro, Malfa, Wright, Li, Shadbolt, Wooldridge, and
  Torr}]{agora}
Samuele Marro, Emanuele~La Malfa, Jesse Wright, Guohao Li, Nigel Shadbolt,
  Michael Wooldridge, and Philip Torr. 2024.
\newblock \href {https://arxiv.org/abs/2410.11905} {A scalable communication
  protocol for networks of large language models}.
\newblock \emph{Preprint}, arXiv:2410.11905.

\bibitem[{Pan et~al.(2024)Pan, Wu, Jiang, Xia, Luo, Zhang, Lin, Rühle, Yang,
  Lin, Zhao, Qiu, and Zhang}]{llmlingua-2}
Zhuoshi Pan, Qianhui Wu, Huiqiang Jiang, Menglin Xia, Xufang Luo, Jue Zhang,
  Qingwei Lin, Victor Rühle, Yuqing Yang, Chin-Yew Lin, H.~Vicky Zhao, Lili
  Qiu, and Dongmei Zhang. 2024.
\newblock \href {https://arxiv.org/abs/2403.12968} {Llmlingua-2: Data
  distillation for efficient and faithful task-agnostic prompt compression}.
\newblock \emph{Preprint}, arXiv:2403.12968.

\bibitem[{Shawn(2026)}]{faithful-not-corrective}
Zayx Shawn. 2026.
\newblock \href {https://arxiv.org/abs/2607.09678} {Faithful, not corrective:
  Message-format effects in multi-hop agent relays are tier-dependent}.
\newblock \emph{Preprint}, arXiv:2607.09678.

\bibitem[{Song(2026)}]{cache-aware-compression}
Yan Song. 2026.
\newblock \href {https://arxiv.org/abs/2607.15516} {Cache-aware prompt
  compression:a two-tier cost model for llm api caching}.
\newblock \emph{Preprint}, arXiv:2607.15516.

\bibitem[{{TensorZero}(2026)}]{tensorzero}
{TensorZero}. 2026.
\newblock Stop comparing price per million tokens: the hidden {LLM} {API}
  costs.
\newblock
  \url{https://www.tensorzero.com/blog/stop-comparing-price-per-million-tokens-the-hidden-llm-api-costs/}.
\newblock Engineering blog post.

\bibitem[{Yao et~al.(2026)Yao, Zou, and Hawkins}]{talk-is-cheap}
Yiheng Yao, Chelsea Zou, and Robert~D. Hawkins. 2026.
\newblock \href {https://arxiv.org/abs/2605.01750} {Talk is cheap,
  communication is hard: Dynamic grounding failures and repair in multi-agent
  negotiation}.
\newblock \emph{Preprint}, arXiv:2605.01750.

\bibitem[{Zhang et~al.(2026)Zhang, Zhang, Pappu, El, Blanchet, Athey, Liu, and
  Zou}]{terms-bench}
Erica Zhang, Fangzhao Zhang, Aneesh Pappu, Batu El, Jose Blanchet, Susan Athey,
  Jiashuo Liu, and James Zou. 2026.
\newblock \href {https://arxiv.org/abs/2605.13909} {Terms-bench: Diagnosing llm
  negotiation agents beyond deal rate}.
\newblock \emph{Preprint}, arXiv:2605.13909.

\end{thebibliography}

\appendix

\openappendixcolumns

\hypertarget{app:related}{%
\section{Extended Related Work}\label{app:related}}

Section 2 keeps the results this study uses as premises. This appendix records the adjacent work that sits on a different axis, and why each one does not answer the acceptance question.

\textbf{Relay fidelity.} Faithful, Not Corrective \citep{faithful-not-corrective} compares constrained, free, JSON, key-value, and triple formats in multi-hop relay. Its effects depend on model tier. The outcome is fidelity rather than cost. The tier dependence points in the same direction as our asymmetric-protocol hypothesis but does not test it.

\textbf{Fixed-protocol evaluation.} ProtocolBench \citep{protocolbench} compares protocols by success, latency, bytes, and resilience. Protocol choice is fixed. It does not model whether a party should accept a proposal.

\textbf{Protocol complexity.} The MCP versus A2A study \citep{mcp-vs-a2a} reports complexity-dependent protocol crossing and already uses the term ``crossover.'' We therefore avoid that term as a headline. Its axis is protocol complexity, not bilateral adoption under private costs.

\textbf{Communication pricing.} Communication Pricing \citep{communication-pricing} combines coalition formation with communication-edge costs. A single designer chooses whom to contact. Our setting has separate decision makers who can disagree about adoption.

\textbf{Negotiation environments.} TERMS-Bench \citep{terms-bench} and Talk is Cheap \citep{talk-is-cheap} provide negotiation tasks with verifiable outcomes. We originally pinned the unrun C0-C5 overlay to Talk is Cheap rather than introduce another negotiation benchmark. That structural choice is recorded in Section 5.1 and is not withdrawn. Appendix \ref{app:suites} records that neither suite had a confirmed Apache 2.0 or MIT source tree. The overlay therefore uses \texttt{negoenv} as a substrate. \texttt{negoenv} is not a contribution of this paper.

\textbf{Post-processing compression.} LLM-Lingua-2 \citep{llmlingua-2} is a non-protocol alternative. Prior agent-format evaluation reports complete task failure beyond 30\% compression. We treat it as a candidate baseline rather than assume that a learned compressed prompt preserves an agent protocol. It is not in the condition registry and this overlay does not run it.

\textbf{Acceptance primitives.} ACP \citep{acp} supplies propose, accept, reject, and counter operations. Its development merged into A2A in 2025 \citep{acp-a2a-merge}, and the primitives carried over. Neither specifies when cost makes rejection rational. That missing criterion is what Section 8 approaches, with measured cost arithmetic rather than the policy comparison the design called for, which the reduced overlay cannot support.

\hypertarget{app:costdetail}{%
\section{Cost Model and Design Details}\label{app:costdetail}}

Section 4 states the bilateral cost model and Section 5 the design. This appendix carries the two parts of the model that the body only names, the sweeps and decision criteria the reduced overlay does not exercise, the per-dialogue record layout, and the list of departures from the two preregistrations.

\hypertarget{why-an-expired-cache-is-not-a-substitute}{%
\subsection{Why an expired cache is not a substitute}\label{why-an-expired-cache-is-not-a-substitute}}

Cache TTL is a vendor product, not a study-wide constant. Cross-organization turns (email, ticket queues, human approval) routinely sit in hours to days. That interval exceeds every \textbf{numeric} maximum TTL documented for the study endpoints. After expiry the realized hit indicator is zero regardless of the nominal \(h_i\). Compression can still reduce \(\mathrm{new}_{i,t}\); the expired cache cannot. The cache-horizon term is therefore treated like negotiation-failure probability \texttt{q}: a parameter sweep, not a missing observational dataset.

Anthropic is the only study vendor for whom both a write premium and a numeric TTL are official on the default OpenRouter-compatible product. At the 1-hour ceiling the sunk write is \$1.00 per million prefix tokens with zero subsequent read discount. Organizational latency above one hour makes that ceiling the relevant bound: paying 2× to write a prefix that expires before the next message is a pure transfer to the vendor. Per-vendor documented TTL and the full write-without-read loss table are in Appendix \ref{app:cache}.

Let \texttt{e} be the share of turns whose interarrival exceeds the purchasable TTL where a numeric TTL exists. The realized hit rate is \(h_i (1 - e)\). We define the sweep of \texttt{e} over \texttt{\{0,\ 0.25,\ 0.5,\ 0.75,\ 1\}} jointly with the nominal hit-rate sweep, which admits sensitivity curves at both extremes: \texttt{e\ =\ 1} is the expired cache and \texttt{e\ =\ 0} the nominal one. This overlay does not report those curves, because the headline pair has no numeric TTL to exceed (Section 8.3, Appendix \ref{app:cache}). No interarrival file is required for this claim. Vendors without a numeric TTL stay at \texttt{e}-insensitive \texttt{h\ =\ 0} reporting plus the hit sweep only when \texttt{cached\_read\_per\_1m} is known.

\hypertarget{the-context-aware-break-even-length}{%
\subsection{The context-aware break-even length}\label{the-context-aware-break-even-length}}

The context-aware version does not force a constant turn saving. Write \(C_{\mathrm{comp},i}(t, h_i)\) and \(C_{\mathrm{eng},i}(t, h_i)\) for party \texttt{i}'s turn-\texttt{t} cost under the compressed and the English representation:

\begin{equation*}
\begin{aligned}
N^{*}_{\mathrm{ctx},i} = {}& \min\, N \ \text{such that} \\
& C_{\mathrm{neg},i} + \sum_{t \le N} \bigl( C_{\mathrm{comp},i}(t, h_i) \\
& \qquad {}- C_{\mathrm{eng},i}(t, h_i) \bigr) < 0.
\end{aligned}
\end{equation*}

This form captures growing histories, prefix caching, and format-dependent tokenization. Mutual adoption at horizon \texttt{N} requires both cumulative differences to be negative. The mutual-benefit set is empty when no tested protocol and horizon satisfy both inequalities.

\hypertarget{cache-activation-and-wire-compression-are-not-substitutes}{%
\subsection{Cache activation and wire compression are not substitutes}\label{cache-activation-and-wire-compression-are-not-substitutes}}

The cache model adds a time axis. A synchronous loop can repeatedly read a stable prefix. An asynchronous exchange can pay to write the prefix and return after expiration. In that region, cache activation and wire compression are not substitutes. They affect different token components and can have opposite party-specific incentives.

\hypertarget{currency-and-token-accounting}{%
\subsection{Currency and token accounting}\label{currency-and-token-accounting}}

Token counts come from each party's native tokenizer. Prices are applied only after tokenization. We report tokens, listed-price cost, cache-adjusted cost, negotiation cost, and repair cost separately. A monetary total without this decomposition is insufficient evidence for bilateral adoption.

\end{multicols}
\begin{appendixwide}
\hypertarget{the-primary-record}{%
\subsection{The primary record}\label{the-primary-record}}

The primary record keeps the parties separate:
\appendixwidetable
\scriptsize
\captionof{table}{The primary per-dialogue record keeps each party's cost fields separate. Party A and party B each carry the seven cost fields, and shared fields record the joint outcome.}\label{tab:record}
\begin{tabularx}{\textwidth}{@{}l>{\hsize=1.000\hsize\RaggedRight\arraybackslash}X@{}}
\toprule
\textbf{Scope} & \textbf{Recorded fields} \\
\midrule
party A & input tokens, output tokens, cache read, cache write, negotiation cost, repair cost (null where compliance is undefined), total cost \\
party B & the same seven fields, measured in B's environment \\
shared & terminal success, outcome equivalence, turns, protocol state \\
\bottomrule
\end{tabularx}
\end{appendixwide}
\begin{multicols}{2}

\hypertarget{cache-and-horizon-sweeps}{%
\subsection{Cache and horizon sweeps}\label{cache-and-horizon-sweeps}}

Cache-hit probability is swept over \texttt{\{0,\ 0.5,\ 0.75,\ 0.9,\ 0.99\}} independently for both parties, and the reported tables carry the no-cache and near-perfect-cache ends of that sweep. The design also defines a sweep of TTL exceedance \texttt{e} over \texttt{\{0,\ 0.25,\ 0.5,\ 0.75,\ 1\}}, in the same way as negotiation-failure probability \texttt{q}. Neither the \texttt{e} curves nor the \texttt{q} curves are reported here: the headline pair documents no numeric TTL to exceed (Appendix \ref{app:cache}), and the negotiation-failure categories were never logged (Appendix \ref{app:negofail}). Break-even is evaluated over observed conversation horizons.

\hypertarget{decision-criteria}{%
\subsection{Decision criteria}\label{decision-criteria}}

The main confirmatory question is whether a tested configuration has no protocol with positive net benefit for both parties. The asymmetric comparison tests whether allowing direction-specific representations expands the mutual-benefit set while preserving the terminal outcome.

The switching policy receives only observable signals available before acceptance. Candidate signals include message length, observed notation family, stated model family when disclosed, prior compliance, and elapsed time since the last exchange. It does not receive the other party's private bill.

\hypertarget{analysis-integrity}{%
\subsection{Analysis integrity}\label{analysis-integrity}}

All result tables are generated from append-only records. P0 is not reused to set a favorable P0 threshold. P0b used a new registration, Amendment 1 before measurement, and a within-family selection-confirmation split. Selection-half family values are not cited. Missing main-experiment records remain unmeasured in Sections 7 and 8 rather than being inferred from P0 or P0b token counts. P0b \texttt{A}, \texttt{S}, and \texttt{K} are token-level statistics. They are not conversation costs.

The P1 dialogue grid is flattened over \texttt{(pair,\ condition,\ difficulty,\ instance,\ seed)}, shuffled with a recorded seed, and executed in that order so that an abort cannot truncate whole pairs and leave a systematically biased remainder.

\hypertarget{deviations-from-preregistration}{%
\subsection{Deviations from preregistration}\label{deviations-from-preregistration}}

The registrations are not edited after the fact. Every departure from them is listed here and repeated where it bears on a result.

\textbf{The P0 gate failed and is reported as a failure.} Compressed dispersion reached 1.280 against a registered 1.300. The threshold was not moved and no passing subset was selected (Section 6.1).

\textbf{P0b is a separate registration, amended before measurement.} Amendment 1 fixed the item count and the selection-confirmation split before any call was issued (Section 6.2). The post-hoc digit-versus-symbol contrast in Section 6.4 is unregistered and is labelled as such.

\textbf{C1 was implemented incorrectly and re-run.} The first overlay injected the canonical schema and switched on a turn counter, so it never negotiated. It is retained as a labelled contrast rather than deleted, and C1 was re-run on 2026-08-25 with handshake verification and a propose-accept-reject loop (Sections 5.2 and 7.1, and Appendix \ref{app:overlay}).

\textbf{C2, C4, and C5 were not run.} The recoupment overlay is C0, C1, and C3 only. Question (iii) and the handshake net-cost contrasts are unmeasured, and the planned switching-policy comparison is replaced by interpretive cost arithmetic (Section 8, Limitations).

\textbf{P1 has no registered gate.} Its cutoffs are design choices, so no P1 number in this paper is a preregistered threshold.

\textbf{The study directory moved from 008 to 005} during consolidation, and the title is set in Title Case wherever it appears. Neither changes a claim, and the earlier registration text keeps its original numbering.

\hypertarget{prompt-construction}{%
\subsection{Prompt construction}\label{prompt-construction}}

Each user message keeps a frozen prefix (instance, rules, schema) and appends a one-line terminal reminder, including turns remaining, only at the end. English turns remind \texttt{ACCEPT} / \texttt{IMPASSE} / \texttt{CONFIRM\ IMPASSE}. Compressed turns remind \texttt{!} / \texttt{\textasciitilde{}} / \ttsplit{\textasciitilde{}confirm}. The two reminders carry the same slots so the English baseline is not given extra strategy. The suffix is required because compressed conditions already re-expose the schema every turn; without it, C0 and C5 forget the terminal tokens. The first reminder is frozen in the cached prefix so consecutive same-party prompts share the previous full user message, and the last line restates the live remaining count. No few-shot examples are added.

Mistral endpoints are excluded from the overlay for two reasons. In the interrupted shuffled 6x6 run, Mistral 429s drove adaptive pacing to 0.1 rps and became the wall-clock bottleneck. In pilot v4, C3 party-B compressed-wire compliance for mistral-small was 0.111, so its compressed cells would not yield a cost decomposition. Both observations are kept in Appendix \ref{app:screening}.

\hypertarget{outcome-classes}{%
\subsection{Outcome classes}\label{outcome-classes}}

\texttt{correct} is an agreement inside F, or an impasse when F is empty. \texttt{search\_failure} is an impasse or \texttt{no\_settlement} when F is nonempty, and also an empty-F timeout with no deal. That class is exploration under a turn budget, not a protocol error. \texttt{error\_empty} is an empty-F agreement (\texttt{false\_agreement}). \ttsplit{error\_below\_reservation} is an agreement whose allocation lies outside a nonempty F, and it is the more general wrong-deal mode. Pilot success cutoffs \texttt{error\ \textless{}\ 0.15} and empty-instance \texttt{correct\ \textgreater{}\ 0.4} are arbitrary and are not preregistered.

\hypertarget{app:suites}{%
\section{Suite Availability}\label{app:suites}}

This appendix records the license and source check for two candidate negotiation suites. The check is a reproducibility fact, not a judgment of the papers. We did not send author mail.

\hypertarget{decision-rule}{%
\subsection{Decision rule}\label{decision-rule}}

Code and data were eligible for vendoring only if both were Apache 2.0 or MIT. CC BY-SA, GPL-family licenses, non-commercial clauses, ``research only'' terms, a missing \texttt{LICENSE}, or an unreachable source tree were treated as ineligible. Ambiguity was treated as ineligible.

\hypertarget{terms-bench-arxiv2605.13909}{%
\subsection{TERMS-Bench (arXiv:2605.13909)}\label{terms-bench-arxiv2605.13909}}

The HTML record lists a code URL \ttsplit{github.com/zou-group/terms-bench} and a project hub \ttsplit{github.com/Terms-bench}. On 2026-08-24, \texttt{GET} \ttsplit{https://github.com/zou-group/terms-bench} returned HTTP 404. The GitHub API for that repository returned \texttt{"Not\ Found"}.

The organization \ttsplit{https://github.com/Terms-bench} had one public repository: \ttsplit{https://github.com/Terms-bench/terms-bench.github.io}, default branch \texttt{main}, commit \ttsplit{aedba60614abb2a312589524435b31f5db025edd} (2026-06-13), GitHub \texttt{license} field null, no \texttt{LICENSE} file. The site \ttsplit{https://terms-bench.github.io/} is a leaderboard with JavaScript result tables, traces, and a system-prompt file. It is not a runnable evaluation loop.

The arXiv abstract license for 2605.13909 is CC BY 4.0. The paper's data-grounded instantiation reuses AmazonHistoryPrice under Apache-2.0 from a third-party repository. That license does not cover TERMS-Bench's own code or generated episodes.

The paper specifies a counterpart kernel and oracle policy in mathematics. Those objects were not available as an Apache 2.0 or MIT source tree. Verdict: not integrable.

\hypertarget{talk-is-cheap-arxiv2605.01750v2}{%
\subsection{Talk is Cheap (arXiv:2605.01750v2)}\label{talk-is-cheap-arxiv2605.01750v2}}

The paper describes an iterated resource-allocation game with verifiable joint optima. Section 4 states that traces are stored and released. Appendix H points to the explorer \ttsplit{https://devyaoyh.github.io/a2a-negotiation/}. The explorer page has no repository URL, no contact address, and no license. It is a SQL interface over embedded traces.

There is no Code Availability section. Appendix B mentions internal git commit hashes on experiment metadata. The arXiv abstract license is CC BY 4.0. GitHub user \texttt{devYaoYH} matches the explorer hostname; no public repository for this paper was listed. \ttsplit{TheNormativityLab/talk-aint-cheap} belongs to a different paper (arXiv:2509.05396) and was not used.

An unsent inquiry draft is at \ttsplit{docs/outreach/talk-is-cheap-license-inquiry.md}. Verdict: not integrable.

\hypertarget{consequence}{%
\subsection{Consequence}\label{consequence}}

Neither suite cleared the decision rule. The C0-C5 overlay therefore uses \texttt{negoenv}, a minimal exhaustive-enumeration environment with an Apache 2.0 file in-tree. \texttt{negoenv} is a substrate. It is not a contribution of this paper. C0-C5 are unchanged condition definitions on top of that substrate.

\hypertarget{app:screening}{%
\section{Unregistered Model Screening}\label{app:screening}}

This appendix reports an unregistered feasibility screen, not a preregistered gate and not a result of the main overlay. The screen asks whether a candidate endpoint can keep the canonical compressed wire when it is the sender. The task is C3 (oracle schema), \texttt{wide} instances 0--2, at most 10 turns, self-play. The pass cutoff is compressed-wire compliance 0.5. That cutoff is arbitrary. A model with compliance 0 would turn every compressed cell in its pairs into a failure observation rather than a cost decomposition. Failures, if any, are kept here rather than dropped.

Source: \ttsplit{results/p1/model\_screening.md}.

\end{multicols}
\begin{appendixwide}
\hypertarget{per-model-compressed-wire-compliance}{%
\subsection{Per-model compressed-wire compliance}\label{per-model-compressed-wire-compliance}}
\appendixwidetable
\footnotesize
\captionof{table}{Per-model C3 self-play compressed-wire compliance in the unregistered screen, with the pass decision at the arbitrary 0.5 cutoff.}\label{tab:d-screen}
\begin{tabular}{@{}llrl@{}}
\toprule
\textbf{Model} & \textbf{Vendor} & \textbf{C3 compliance} & \textbf{Pass} \\
\midrule
\ttsplit{openai/gpt-5.4-nano} & openai & 1.000 & yes \\
\ttsplit{google/gemini-2.5-flash} & google & 0.967 & yes \\
\ttsplit{anthropic/claude-haiku-4.5} & anthropic & 1.000 & yes \\
\ttsplit{qwen/qwen3.7-flash} & alibaba & 1.000 & yes \\
\ttsplit{deepseek/deepseek-chat-v3.1} & deepseek & 1.000 & yes \\
\ttsplit{mistralai/mistral-small-3.2-24b-instruct} & mistral & 0.544 & yes \\
\bottomrule
\end{tabular}
\end{appendixwide}
\begin{multicols}{2}

Pass count: 6 of 6. Threshold 0.5 (arbitrary).

Mistral-small's three instance rates were 0.333, 0.800, and 0.500. The pooled 0.544 clears the cutoff. The same model is the cheapest remaining cross-vendor partner for Qwen on the P0b price snapshot, so it remains in the cheap pilot pair. That is a screen pass, not evidence that the model holds the protocol under C1 or C2.

\hypertarget{small-models-and-protocol-maintenance}{%
\subsection{Small models and protocol maintenance}\label{small-models-and-protocol-maintenance}}

The screen is a sender-side C3 self-play. Five of six endpoints stay on the schema on almost every compressed turn. The remaining endpoint, a smaller instruction model, is near the cutoff and drops below 0.5 on one of three instances. This is a screening observation: some smaller models do not reliably emit the compressed wire even when the schema is injected and both roles are the same model. It is not a confirmatory finding about capability in the main experiment, and it is not a registered test of (ii) or (iii).

\hypertarget{implications-for-ii-and-iii-screening-strength-only}{%
\subsection{Implications for (ii) and (iii), screening strength only}\label{implications-for-ii-and-iii-screening-strength-only}}

Claim (ii) (Section 1) is token-level evidence that a representation can change who is relatively cheaper; an empty mutual-benefit set at conversation cost still requires P2. Claim (iii) is whether direction-specific representations expand that set, which is not yet measured because C2 was not run. Party-specific break-even under cache and horizon is claim (iv), addressed at billed cost in Sections 7 and 8. If one party cannot stay on the compressed wire, billed tokens mix tokenizer effects with repair and parse failure. That mixture can look like an empty mutual-benefit set or a shifted break-even even when prices and cache are unchanged. This argument applies to conditions whose grammar is fixed and checkable, which in the measured overlay is C3 alone: C0 has no compressed turn, and implemented C1 negotiates its grammar per dialogue, so neither yields a violation rate to reason from (Limitations). The screen therefore treats protocol competence as a prerequisite for reading a compressed condition as cost. The 0.5 cutoff and the C3 self-play task are design choices, not evidence that (ii) or (iii) hold or fail. Main-experiment cells remain unrun for C2, C4, and C5. The later recoupment overlay uses C0, C1, and C3.

\hypertarget{cheap-pair-pilot-retained-excluded-from-the-overlay}{%
\subsection{Cheap-pair pilot retained; excluded from the overlay}\label{cheap-pair-pilot-retained-excluded-from-the-overlay}}

\ttsplit{mistralai/mistral-small-3.2-24b-instruct} stays in the pilot record and this appendix. C3 self-play screening compliance was 0.544. In the v4 two-party C3 dialogues the same model as party B had compressed-wire compliance 0.111. Oracle schema in isolation is held about half the time; the same endpoint drops the schema in a multi-turn bargain. That is a direct observation of multi-turn drift, in the same direction as Multi-IF instruction decay. It is usable material for claim (iii) (asymmetric protocols) because B's failure is party-specific. It is a screening-plus-pilot observation, not a confirmatory result.

Cheap-pair compressed numbers are pilot observations. They are not headline summaries and are not merged into the 1053-dialogue overlay.

\hypertarget{overlay-exclusion-observations-kept}{%
\subsection{Overlay exclusion (observations kept)}\label{overlay-exclusion-observations-kept}}

\ttsplit{mistralai/mistral-small-3.2-24b-instruct} remains in this appendix. It is dropped from the 1215-dialogue C0/C1/C3 overlay only. Two operational facts, not a change to the 0.5 screen: (1) in the interrupted shuffled 6×6 run, Mistral 429s adapted rps to 0.1 and gated the whole matrix; (2) v4 C3-B compressed-wire compliance was 0.111, so compressed cells on that pair are not a usable cost series. D.4 is unchanged.

\hypertarget{app:cache}{%
\section{Vendor Cache Pricing and TTL}\label{app:cache}}

Per-vendor documented cache TTL and cache-write pricing referenced by Section 8.3 and Appendix \ref{app:costdetail}. Prices, TTLs, and retrieval URLs are snapshotted in \ttsplit{data/design/cost\_parameters.json} (verified 2026-08-24). No API calls.

\end{multicols}
\begin{appendixwide}
\hypertarget{documented-ttl-by-vendor-verified-2026-08-24}{%
\subsection{Documented TTL by vendor (verified 2026-08-24)}\label{documented-ttl-by-vendor-verified-2026-08-24}}
\appendixwidetable
\scriptsize
\captionof{table}{Documented cache mode, TTL, write billing, and storage price per study vendor, verified 2026-08-24.}\label{tab:e-ttl}
\begin{tabularx}{\textwidth}{@{}l>{\hsize=0.695\hsize\RaggedRight\arraybackslash}X>{\hsize=0.805\hsize\RaggedRight\arraybackslash}X>{\hsize=1.079\hsize\RaggedRight\arraybackslash}X>{\hsize=1.134\hsize\RaggedRight\arraybackslash}X>{\hsize=0.823\hsize\RaggedRight\arraybackslash}X>{\hsize=1.463\hsize\RaggedRight\arraybackslash}X@{}}
\toprule
\textbf{Vendor} & \textbf{Endpoint} & \textbf{Caching mode} & \textbf{Default TTL} & \textbf{Maximum documented TTL} & \textbf{Write billing} & \textbf{Storage} \\
\midrule
OpenAI & \ttsplit{gpt-5.4-nano} & automatic & not a single product default (typical 5--10 min inactivity) & in-memory ``up to about 1 hour''; 24 h list does not name nano & write = uncached input (1.0×) for pre-GPT-5.6 & none (in-memory) \\
Google & \ttsplit{gemini-2.5-flash} & automatic implicit (OpenRouter path) & not numeric & explicit caches default 1 h and have no published max & implicit create = 100\% input (1.0×) & implicit: \$0 / 1M / h (documented free). Explicit product: \$1.00 / 1M tokens / h \\
Anthropic & \ttsplit{claude-haiku-4.5} & explicit \ttsplit{cache\_control} & 5 min & 1 h (\ttsplit{ttl=1h}) & 1.25× at 5 min; 2× at 1 h & none on the pricing page \\
Alibaba & \ttsplit{qwen3.7-flash} & automatic implicit on Model Studio & indeterminate & explicit option 5 min (not used on this OpenRouter path) & implicit create 100\%; explicit create 125\% & none \\
DeepSeek & \ttsplit{deepseek-chat-v3.1} & automatic (OpenRouter) & not numeric (``hours to days'' in KV-cache docs) & not numeric & \textbf{unverified for V3.1} & unverified \\
Mistral & \ttsplit{mistral-small-3.2-24b-instruct} & automatic with optional \ttsplit{prompt\_cache\_key} & \textbf{not stated} & \textbf{not stated} & \textbf{not stated} & none stated \\
\bottomrule
\end{tabularx}
\end{appendixwide}
\begin{multicols}{2}

OpenAI, Google implicit, Alibaba implicit, DeepSeek, and Mistral are excluded from Table \ref{tab:e-writeloss} whenever the write multiplier or the TTL is null. They are not filled with Anthropic's 1.25× / 5 min figures.

\end{multicols}
\begin{appendixwide}
\hypertarget{app:writeloss}{%
\subsection{Write without a later read}\label{app:writeloss}}

If inter-turn time exceeds TTL, the prefix is written (or stored) and never read. Net loss per 1 million prefix tokens, using only documented numbers:
\appendixwidetable
\scriptsize
\captionof{table}{Net loss per 1 million prefix tokens when a written prefix is never read, using documented write multipliers and TTLs only.}\label{tab:e-writeloss}
\begin{tabularx}{\textwidth}{@{}>{\hsize=0.620\hsize\RaggedRight\arraybackslash}X>{\hsize=1.612\hsize\RaggedRight\arraybackslash}X>{\hsize=0.620\hsize\RaggedRight\arraybackslash}X>{\hsize=0.620\hsize\RaggedRight\arraybackslash}X>{\hsize=1.136\hsize\RaggedRight\arraybackslash}X>{\hsize=0.620\hsize\RaggedRight\arraybackslash}X>{\hsize=1.771\hsize\RaggedRight\arraybackslash}X@{}}
\toprule
\textbf{Vendor} & \textbf{Path} & \textbf{Write multiplier} & \textbf{TTL} & \textbf{Token write premium} & \textbf{Storage over TTL} & \textbf{Net loss USD / 1M prefix tokens} \\
\midrule
Anthropic & default & 1.25× & 5 min & \$0.25 & \$0 & \textbf{\$0.25} \\
Anthropic & extended & 2× & 1 h & \$1.00 & \$0 & \textbf{\$1.00} \\
Google & explicit context cache (not the OpenRouter implicit path) & 1.0× & 1 h default & \$0.00 & \$1.00 & \textbf{\$1.00} \\
Alibaba & explicit (not used here) & 1.25× & 5 min & \$0.0075 at the \$0.03 / 1M listed prompt & \$0 & \textbf{\$0.0075} \\
OpenAI nano & automatic & 1.0× & TTL field null & n/a & n/a & excluded (no numeric TTL; write premium is 0 if a write occurs) \\
Google implicit & OpenRouter & 1.0× & TTL field null & n/a & \$0 & excluded from TTL-expiry accounting \\
DeepSeek V3.1 &  & null & null &  &  & excluded \\
Mistral &  & null & null &  &  & excluded \\
\bottomrule
\end{tabularx}
\end{appendixwide}
\begin{multicols}{2}

\hypertarget{asynchronous-write-loss}{%
\subsection{Asynchronous write loss}\label{asynchronous-write-loss}}

Long inter-turn delays can remove the read that justifies a cache write. Table \ref{tab:e-writeloss} above gives the full loss. Among the study endpoints a numeric default-path premium is documented only for Anthropic.

The headline pair uses OpenAI and Google implicit caching. Neither endpoint has both a numeric TTL and a write premium in the parameter file. We therefore do not transfer Anthropic's loss to that pair. The explicit Google and Alibaba products listed above were not used by the runner.

\hypertarget{app:p0gate}{%
\section{The P0 Gate in Full}\label{app:p0gate}}

Section 6.1 records that the P0 gate failed. This appendix carries the counting method, why the registered statistic did not operationalize the claim it was meant to screen, the descriptive observations that followed the failure, and the provenance of the notation-family values. Every statistic in F.3 and F.4 was selected after seeing P0 and is post hoc.

\end{multicols}
\begin{appendixwide}
\hypertarget{data-and-counting}{%
\subsection{Data and counting}\label{data-and-counting}}

P0 uses the P0 item set from the source experiment. It contains 100 English messages and 100 content-matched compressed messages. We count each surface with native token accounting from one model in each of 6 vendor families. This yields 1,200 complete cells and no missing cells.

Counts are \texttt{usage.prompt\_tokens} returned by the provider through OpenRouter. Generation is capped at 1 token. No character-based estimate enters the analysis.
\appendixwidetable
\footnotesize
\captionof{table}{P0 native prompt-token totals over all 100 items, by vendor, English against compressed.}\label{tab:p0-vendors}
\begin{tabular}{@{}lrrr@{}}
\toprule
\textbf{Vendor} & \textbf{English} & \textbf{Compressed} & \textbf{Compressed / English} \\
\midrule
OpenAI & 20,910 & 17,270 & 0.826 \\
DeepSeek & 21,106 & 16,377 & 0.776 \\
Mistral & 23,350 & 20,441 & 0.875 \\
Alibaba & 23,790 & 19,402 & 0.816 \\
Google & 23,800 & 18,776 & 0.789 \\
Anthropic & 24,850 & 20,962 & 0.844 \\
\bottomrule
\end{tabular}
\end{appendixwide}
\begin{multicols}{2}

\hypertarget{why-the-gate-did-not-test-the-empty-set-claim}{%
\subsection{Why the gate did not test the empty-set claim}\label{why-the-gate-did-not-test-the-empty-set-claim}}

The empty-set claim is the possibility that the Section 4.2 mutual-benefit set is empty. It is an existence claim about sign: a configuration exists in which no protocol benefits both parties. The registered gate measured magnitude: maximum divided by minimum token count. These are different propositions.

Large dispersion does not imply the empty-set claim. Every party could benefit with different magnitudes. Small dispersion does not refute the empty-set claim. Two parties can have nearly equal magnitudes with opposite signs. P0 therefore did not operationalize the claim it was intended to screen.

The maximum-to-minimum statistic has another problem. It depends on which vendors enter the list. Vendor selection is a researcher degree of freedom. That dependence is methodologically more important than the 1.5\% relative shortfall between 1.280 and 1.300. The threshold of 1.300 also had no independent empirical basis. The correct response is a new registration, not a reinterpretation of this gate.

\hypertarget{descriptive-observations-after-the-failed-gate}{%
\subsection{Descriptive observations after the failed gate}\label{descriptive-observations-after-the-failed-gate}}

The following statistics were not preregistered. We label them post hoc.

\textbf{Party-specific benefit spread.} The same compressed set reduces DeepSeek tokens by 22.4\% and Mistral tokens by 12.5\%. The ratio of their retained-token fractions is \texttt{0.875\ /\ 0.776\ =\ 1.128}. This paired contrast is the most direct P0 evidence for ``cheaper for whom.'' Each vendor sees the same semantic payload in English and compressed form, so general differences in verbosity or vocabulary size do not by themselves create the within-vendor saving.

\textbf{Dispersion amplification.} English maximum-to-minimum dispersion is 1.188. Compressed dispersion is 1.280. Their ratio indicates a 7.7\% increase in cross-vendor dispersion under compression. A ratio of spreads is still sensitive to endpoint membership, but it compares the same vendor list on both surfaces and is less exposed than an isolated extreme-value threshold.

\textbf{Rank reversal.} The 4 reversing pairs are Alibaba-Google, Alibaba-Mistral, DeepSeek-OpenAI, and Google-Mistral. This is evidence for claim (ii) because a representation changes who is relatively cheaper. It is not a test of the empty-set claim. The outcome is token count, not net conversation cost, and contains no handshake, violation, repair, price, or horizon term.

\hypertarget{provenance-of-notation-family-observations}{%
\subsection{Provenance of notation-family observations}\label{provenance-of-notation-family-observations}}

Round 1 mixed JSON, YAML, TOON, TSV, URL-query, and symbol-dense forms. Its pooled compressed dispersion was 1.174. The \texttt{unicode\_ops} family alone reached 1.385. We observed that value before replacing the Round 1 stimulus.

Round 2 uses symbol-dense forms. Its pooled dispersion is 1.280. Five families exceed 1.300: \texttt{set\_builder} at 1.562, \texttt{unicode\_ops} at 1.550, \texttt{rpn} at 1.407, \texttt{apl\_arrow} at 1.360, and \texttt{pipe\_kv} at 1.348.

Selecting those five after seeing Round 2 would be winner's curse, not a rescue of the registered test. We report four of them as exploratory. \texttt{unicode\_ops} has a different provenance. It was selected by its Round 1 value before the Round 2 redesign, so the Round 2 value of 1.550 is an out-of-sample confirmation for that family. It does not convert the pooled gate into a pass.

\hypertarget{app:p0tables}{%
\section{P0 and P0b Tables}\label{app:p0tables}}

Family-by-vendor cells for P0 come from native counts on \texttt{data/p0/items.jsonl}, summarized in \ttsplit{results/p0/summary.json} (Round 2) and \ttsplit{results/p0/round1/summary.json} (Round 1). P0b tables come from \ttsplit{data/p0b/raw\_usage.jsonl} and \ttsplit{results/p0b/VERDICT.md}. Confirmation-half family rows follow \ttsplit{results/p0b/families.md}. Selection-half family values are not cited.

\hypertarget{round-2-vendor-totals}{%
\subsection{Round 2 vendor totals}\label{round-2-vendor-totals}}

Table \ref{tab:p0-vendors} in Appendix \ref{app:p0gate} already carries these six rows, so they are not reprinted. Source: \ttsplit{results/p0/summary.json}, generated by \texttt{src/p0\_analyze.py}.

\end{multicols}
\begin{appendixwide}
\hypertarget{aggregate-contrast-and-rank-reversals}{%
\subsection{Aggregate contrast and rank reversals}\label{aggregate-contrast-and-rank-reversals}}
\appendixwidetable
\centering
\footnotesize
\captionof{table}{Round 2 aggregate maximum-to-minimum dispersion, English against compressed.}\label{tab:b-r2-dispersion}
\begin{tabular}{@{}lr@{}}
\toprule
\textbf{Surface} & \textbf{Maximum / minimum} \\
\midrule
English & 1.188 \\
Compressed & 1.280 \\
\bottomrule
\end{tabular}

\appendixwidegap
\footnotesize
\captionof{table}{Round 2 pairs whose cheaper endpoint reverses between surfaces.}\label{tab:b-r2-reversals}
\begin{tabular}{@{}lll@{}}
\toprule
\textbf{Pair} & \textbf{Cheaper in English} & \textbf{Cheaper when compressed} \\
\midrule
Alibaba / Google & Alibaba & Google \\
Alibaba / Mistral & Mistral & Alibaba \\
DeepSeek / OpenAI & OpenAI & DeepSeek \\
Google / Mistral & Mistral & Google \\
\bottomrule
\end{tabular}
\end{appendixwide}
\begin{multicols}{2}

Source: \ttsplit{results/p0/summary.json}, generated by \texttt{src/p0\_analyze.py}.

\end{multicols}
\begin{appendixwide}
\hypertarget{family-level-dispersion}{%
\subsection{Family-level dispersion}\label{family-level-dispersion}}
\appendixwidetable
\footnotesize
\captionof{table}{Round 2 family-level compressed dispersion with provenance.}\label{tab:b-family-dispersion}
\begin{tabular}{@{}lrl@{}}
\toprule
\textbf{Family} & \textbf{Compressed maximum / minimum} & \textbf{Provenance} \\
\midrule
\ttsplit{set\_builder} & 1.562 & Round 2 exploratory \\
\ttsplit{unicode\_ops} & 1.550 & Round 1 selection, Round 2 confirmation \\
\ttsplit{rpn} & 1.407 & Round 2 exploratory \\
\ttsplit{apl\_arrow} & 1.360 & Round 2 exploratory \\
\ttsplit{pipe\_kv} & 1.348 & Round 2 exploratory \\
\bottomrule
\end{tabular}
\end{appendixwide}
\begin{multicols}{2}

\end{multicols}
\begin{appendixwide}
\hypertarget{round-2-family-by-vendor-compressed-tokens}{%
\subsection{Round 2 family-by-vendor compressed tokens}\label{round-2-family-by-vendor-compressed-tokens}}
\appendixwidetable
\footnotesize
\captionof{table}{Round 2 compressed tokens by family and vendor.}\label{tab:b-r2-family}
\begin{tabular}{@{}lrrrrrr@{}}
\toprule
\textbf{Family} & \textbf{OpenAI} & \textbf{Google} & \textbf{Anthropic} & \textbf{Alibaba} & \textbf{DeepSeek} & \textbf{Mistral} \\
\midrule
\ttsplit{pipe\_kv} & 1,936 & 1,947 & 2,153 & 2,108 & 1,742 & 2,348 \\
\ttsplit{sexp} & 1,510 & 1,734 & 1,848 & 1,790 & 1,529 & 1,788 \\
\ttsplit{fieldnum} & 1,530 & 1,887 & 1,721 & 1,879 & 1,567 & 1,865 \\
\ttsplit{unicode\_ops} & 1,884 & 1,886 & 2,381 & 1,874 & 1,536 & 2,139 \\
\ttsplit{agent\_trace} & 1,591 & 1,874 & 1,905 & 1,922 & 1,748 & 2,036 \\
\ttsplit{rpn} & 1,606 & 1,618 & 1,977 & 1,690 & 1,405 & 1,806 \\
\ttsplit{boxed} & 1,823 & 1,728 & 1,955 & 1,943 & 1,684 & 2,082 \\
\ttsplit{set\_builder} & 2,221 & 2,494 & 3,119 & 2,266 & 1,997 & 2,515 \\
\ttsplit{apl\_arrow} & 1,766 & 1,981 & 2,378 & 2,241 & 1,749 & 2,209 \\
\ttsplit{record\_slash} & 1,403 & 1,627 & 1,525 & 1,689 & 1,420 & 1,653 \\
\bottomrule
\end{tabular}
\end{appendixwide}
\begin{multicols}{2}

Source: \ttsplit{results/p0/summary.json}.

\end{multicols}
\begin{appendixwide}
\hypertarget{round-1-contrast}{%
\subsection{Round 1 contrast}\label{round-1-contrast}}
\appendixwidetable
\centering
\footnotesize
\captionof{table}{Round 1 pooled and \texttt{unicode\_ops} compressed dispersion.}\label{tab:b-r1-contrast}
\begin{tabular}{@{}lr@{}}
\toprule
\textbf{Quantity} & \textbf{Value} \\
\midrule
Pooled compressed maximum / minimum & 1.174 \\
\ttsplit{unicode\_ops} maximum / minimum & 1.385 \\
\bottomrule
\end{tabular}
\end{appendixwide}
\begin{multicols}{2}

\end{multicols}
\begin{appendixwide}
\hypertarget{round-1-family-by-vendor-compressed-tokens}{%
\subsection{Round 1 family-by-vendor compressed tokens}\label{round-1-family-by-vendor-compressed-tokens}}
\appendixwidetable
\footnotesize
\captionof{table}{Round 1 compressed tokens by family and vendor.}\label{tab:b-r1-family}
\begin{tabular}{@{}lrrrrrr@{}}
\toprule
\textbf{Family} & \textbf{OpenAI} & \textbf{Google} & \textbf{Anthropic} & \textbf{Alibaba} & \textbf{DeepSeek} & \textbf{Mistral} \\
\midrule
\ttsplit{pipe\_kv} & 966 & 1,067 & 1,090 & 1,109 & 976 & 1,105 \\
\ttsplit{sexp} & 930 & 1,056 & 1,117 & 1,090 & 949 & 1,056 \\
\ttsplit{json\_short} & 970 & 1,060 & 1,092 & 1,115 & 996 & 1,114 \\
\ttsplit{yaml\_flow} & 1,018 & 1,138 & 1,142 & 1,179 & 1,027 & 1,139 \\
\ttsplit{toon} & 1,024 & 1,146 & 1,177 & 1,187 & 1,036 & 1,143 \\
\ttsplit{fieldnum} & 937 & 1,040 & 1,052 & 1,093 & 953 & 1,073 \\
\ttsplit{unicode\_ops} & 987 & 1,070 & 1,295 & 1,090 & 935 & 1,130 \\
\ttsplit{path\_query} & 1,181 & 1,335 & 1,383 & 1,340 & 1,232 & 1,337 \\
\ttsplit{table\_tsv} & 908 & 1,080 & 1,082 & 1,063 & 935 & 1,046 \\
\ttsplit{agent\_trace} & 955 & 1,060 & 1,163 & 1,108 & 1,056 & 1,169 \\
\bottomrule
\end{tabular}
\end{appendixwide}
\begin{multicols}{2}

Source: \ttsplit{results/p0/round1/summary.json}.

\end{multicols}
\begin{appendixwide}
\hypertarget{p0b-pair-conflict-rates}{%
\subsection{P0b pair conflict rates}\label{p0b-pair-conflict-rates}}

Table \ref{tab:b-p0b-pairs} gives the rate for all 15 unordered pairs.

Bootstrap conflict rates for all 15 unordered vendor pairs. Source: \ttsplit{results/p0b/VERDICT.md}.
\appendixwidetable
\centering
\footnotesize
\captionof{table}{P0b bootstrap sign-conflict rate for all 15 vendor pairs.}\label{tab:b-p0b-pairs}
\begin{tabular}{@{}lr@{}}
\toprule
\textbf{Pair} & \textbf{Rate} \\
\midrule
alibaba/anthropic & 0.0 \\
alibaba/deepseek & 0.0 \\
alibaba/google & 0.0 \\
alibaba/mistral & 1.0 \\
alibaba/openai & 0.0 \\
anthropic/deepseek & 0.0 \\
anthropic/google & 1.0 \\
anthropic/mistral & 0.0 \\
anthropic/openai & 0.0 \\
deepseek/google & 0.0 \\
deepseek/mistral & 0.0 \\
deepseek/openai & 0.0 \\
google/mistral & 0.0 \\
google/openai & 0.0 \\
mistral/openai & 0.0 \\
\bottomrule
\end{tabular}
\end{appendixwide}
\begin{multicols}{2}

\end{multicols}
\begin{appendixwide}
\hypertarget{p0b-confirmation-half-family-by-vendor-tokens}{%
\subsection{P0b confirmation-half family-by-vendor tokens}\label{p0b-confirmation-half-family-by-vendor-tokens}}

Tables \ref{tab:b-p0b-eng} and \ref{tab:b-p0b-comp} give the two surfaces.

Confirmation half only. A selection half exists and is not cited. Source: \ttsplit{data/p0b/raw\_usage.jsonl} filtered to \texttt{holdout\ =\ confirmation}. Item counts match \ttsplit{results/p0b/families.md}.
\appendixwidetable
\footnotesize
\captionof{table}{P0b confirmation-half English tokens by family and vendor.}\label{tab:b-p0b-eng}
\begin{tabular}{@{}lrrrrrrr@{}}
\toprule
\textbf{Family} & \textbf{Items} & \textbf{OpenAI} & \textbf{Google} & \textbf{Anthropic} & \textbf{Alibaba} & \textbf{DeepSeek} & \textbf{Mistral} \\
\midrule
\ttsplit{apl\_arrow} & 14 & 2,184 & 2,529 & 2,284 & 2,666 & 2,165 & 2,568 \\
\ttsplit{json\_baseline} & 18 & 2,733 & 3,154 & 2,860 & 3,331 & 2,720 & 3,205 \\
\ttsplit{pipe\_kv} & 12 & 1,972 & 2,297 & 2,061 & 2,414 & 1,954 & 2,330 \\
\ttsplit{rpn} & 16 & 2,446 & 2,822 & 2,558 & 2,978 & 2,438 & 2,866 \\
\ttsplit{set\_builder} & 15 & 2,490 & 2,901 & 2,602 & 3,050 & 2,475 & 2,945 \\
\ttsplit{unicode\_ops} & 17 & 2,927 & 3,393 & 3,060 & 3,560 & 2,900 & 3,441 \\
\bottomrule
\end{tabular}

\appendixwidegap
\footnotesize
\captionof{table}{P0b confirmation-half compressed tokens by family and vendor.}\label{tab:b-p0b-comp}
\begin{tabular}{@{}lrrrrrrr@{}}
\toprule
\textbf{Family} & \textbf{Items} & \textbf{OpenAI} & \textbf{Google} & \textbf{Anthropic} & \textbf{Alibaba} & \textbf{DeepSeek} & \textbf{Mistral} \\
\midrule
\ttsplit{apl\_arrow} & 14 & 2,058 & 2,391 & 2,620 & 2,764 & 2,023 & 2,704 \\
\ttsplit{json\_baseline} & 18 & 3,066 & 3,473 & 3,193 & 3,646 & 3,150 & 3,654 \\
\ttsplit{pipe\_kv} & 12 & 1,507 & 1,850 & 1,701 & 1,936 & 1,515 & 2,000 \\
\ttsplit{rpn} & 16 & 2,098 & 2,164 & 2,476 & 2,334 & 1,793 & 2,462 \\
\ttsplit{set\_builder} & 15 & 2,757 & 2,983 & 2,950 & 3,235 & 2,478 & 3,335 \\
\ttsplit{unicode\_ops} & 17 & 2,893 & 2,961 & 3,484 & 3,125 & 2,372 & 3,549 \\
\bottomrule
\end{tabular}
\end{appendixwide}
\begin{multicols}{2}

\end{multicols}
\begin{appendixwide}
\hypertarget{p0b-registered-gates-and-vendor-totals}{%
\subsection{P0b registered gates and vendor totals}\label{p0b-registered-gates-and-vendor-totals}}

Tables \ref{tab:p0b-gates} and \ref{tab:p0b-vendors} are the gate outcomes and the per-vendor totals Section 6.3 reads.
\appendixwidetable
\footnotesize
\captionof{table}{P0b registered gates, each gate's outcome, and whether it is required, reporting only, or a data-quality check.}\label{tab:p0b-gates}
\begin{tabular}{@{}lll@{}}
\toprule
\textbf{Gate} & \textbf{Result} & \textbf{Role} \\
\midrule
G1 (\ttsplit{A} CI lower \textgreater{} 1.000) & True & required \\
G2 (\ttsplit{S} CI lower \textgreater{} 1.050) & True & reporting only \\
G3 (≥1 pair with conflict rate ≥ 0.95) & True & required \\
Missing rate ≤ 0.05 & True & data quality \\
\bottomrule
\end{tabular}

\appendixwidegap
\footnotesize
\captionof{table}{P0b native prompt-token totals over all 198 items, by vendor, English against compressed.}\label{tab:p0b-vendors}
\begin{tabular}{@{}lrrr@{}}
\toprule
\textbf{Vendor} & \textbf{English} & \textbf{Compressed} & \textbf{Compressed / English} \\
\midrule
DeepSeek & 30,682 & 27,757 & 0.905 \\
OpenAI & 30,888 & 29,766 & 0.964 \\
Anthropic & 32,307 & 34,101 & 1.056 \\
Google & 35,754 & 32,911 & 0.920 \\
Mistral & 36,298 & 36,753 & 1.013 \\
Alibaba & 37,684 & 35,540 & 0.943 \\
\bottomrule
\end{tabular}
\end{appendixwide}
\begin{multicols}{2}

\hypertarget{numeral-density-versus-symbols-in-full}{%
\subsection{Numeral density versus symbols, in full}\label{numeral-density-versus-symbols-in-full}}

This section is an unregistered post-hoc diagnostic. It is not a gate. It does not amend the P0b preregistration (Appendix \ref{app:artifacts}). Section 6.4 states its conclusion; the family-level values are here.

The P0b stimuli are identifier- and numeral-dense (\texttt{ID-0166-01}, \texttt{B50}, \texttt{839}, \texttt{6.1}). Vendors tokenize digits differently. Compression drops English filler (\texttt{Please}, \texttt{has\ quantity}, \texttt{and\ status}) and therefore raises the numeral share. Pooled \texttt{A\ \textgreater{}\ 1} could then come from digit packing rather than from exotic operators.

\texttt{json\_baseline} keeps the same numeric payload without exotic operators. Its confirmation-half \texttt{A} is 0.973 (95\% CI {[}0.967, 0.979{]}). The entire interval lies below 1. Digit-dense JSON without exotic symbols does not amplify cross-vendor spread on this control.

Point estimates of confirmation-half \texttt{A} increase across the five symbol families: \texttt{pipe\_kv} 1.074 (residual 0.101), \texttt{set\_builder} 1.092 (0.119), \texttt{apl\_arrow} 1.110 (0.136), \texttt{rpn} 1.131 (0.157), \texttt{unicode\_ops} 1.219 (0.246).\footnote{Family-level rows use the confirmation half only. A selection half exists and is not cited. \texttt{unicode\_ops} is the registered provenance exception: it reached compressed dispersion 1.385 in Round 1, when stimuli mixed JSON, YAML, and TSV, and was selected before the Round 2 stimulus replacement. Its all-item P0b row in the P0b family estimates (Appendix \ref{app:artifacts}) therefore does not share the out-of-sample status of the other families. The confirmation-half \texttt{unicode\_ops} row used above remains a held-out family estimate.} Adjacent rank is not established. The intervals for \texttt{pipe\_kv} ({[}1.057, 1.087{]}) and \texttt{set\_builder} ({[}1.085, 1.098{]}) overlap on 1.085--1.087. The endpoints are separated: \texttt{pipe\_kv} {[}1.057, 1.087{]} and \texttt{unicode\_ops} {[}1.213, 1.225{]} do not overlap. The argument does not rest on family order. It rests on location: all five symbol families sit above the \texttt{json\_baseline} CI upper bound of 0.979, and \texttt{json\_baseline} itself has its entire interval below 1.

This unregistered post-hoc diagnostic suggests that pooled amplification is not attributable to numerals alone on the JSON control. It does not demonstrate a causal decomposition of \texttt{A} into digit and symbol components. \texttt{json\_baseline} is a lower bound on the digit contribution, not a complete accounting.

Family \texttt{A} values are confirmation-half estimates.

\hypertarget{what-the-family-values-do-and-do-not-show}{%
\subsection{What the family values do and do not show}\label{what-the-family-values-do-and-do-not-show}}

An unregistered post-hoc diagnostic shows confirmation-half point estimates of \texttt{A} rising across the five symbol families, which is consistent with the observation that more exotic operators go with larger amplification. Adjacent rank is not established: \texttt{pipe\_kv} (1.074, CI {[}1.057, 1.087{]}) and \texttt{set\_builder} (1.092, CI {[}1.085, 1.098{]}) overlap on 1.085--1.087. We also did not quantify exoticness independently of \texttt{A}. What the diagnostic supports is location, not order: every symbol family sits above the \texttt{json\_baseline} CI upper bound of 0.979, and \texttt{json\_baseline} itself lies entirely below 1. That pattern suggests numerals alone do not produce pooled \texttt{A\ \textgreater{}\ 1} on this template. It is not a registered decomposition.

The \texttt{unicode\_ops} Round 1 provenance also clarifies what the pooled result means. Symbol-heavy notation can create tokenizer divergence, but the effect is notation-specific. A broad label such as ``compressed protocol'' hides that heterogeneity. Selecting only successful families after observation would overstate it.

\hypertarget{the-p0b-registration-in-full}{%
\subsection{The P0b registration in full}\label{the-p0b-registration-in-full}}

P0b is registered separately and reuses no Round 2 item or count. It splits the two jobs the failed gate confused. The primary statistic is the amplification ratio \(A = D_{\mathrm{comp}} / D_{\mathrm{eng}}\), where each \(D\) is the same-list maximum-to-minimum native prompt-token total. The directional screen is \texttt{K}, the number of unordered vendor pairs whose English token-order sign disagrees with their compressed sign; tied signs are not conflicts. The auxiliary statistic is the benefit-spread \(S = \max_v(r_v) / \min_v(r_v)\) with \(r_v = \mathrm{compressed}_v / \mathrm{english}_v\), which is reporting only. G2 asks whether the interval for \texttt{S} clears 1.050, a cut with no independent prior basis, and it is not part of the pass rule.

\texttt{PASS\ =\ G1\ AND\ G3}. G1 requires the 95\% interval lower bound for \texttt{A} to exceed 1.000, the hypothesis boundary for amplification. G3 requires at least one pair to conflict in at least 95\% of 10,000 paired bootstrap replicates, the existence boundary for sign conflict. Neither bound is an empirical constant taken from P0.

P0b uses 198 new content-matched pairs. Amendment 1, filed before any P0b measurement existed, fixed the count at 198 and listed families alphabetically, because 200 is not divisible by six and the original remainder assignment tracked Round 2 dispersion rank, a result-dependent degree of freedom. Gates, statistics, the holdout salt, and the bootstrap replicate count were unchanged. Item identities split within each family by a fixed hash into selection and confirmation halves, and family-level numbers here use the confirmation half only.

\hypertarget{what-a-single-compression-ratio-hides}{%
\subsection{What a single compression ratio hides}\label{what-a-single-compression-ratio-hides}}

Compression does not act like a uniform multiplier across endpoints, which is the operational form of the result. A sender that estimates savings with its own tokenizer is not estimating the receiver's incentive. The reading is not that one vendor is globally expensive: the same party moves in relative rank when content changes, so a useful switching policy conditions on message and protocol features rather than storing one vendor ordering. Symbol-heavy notation is also not one treatment, and a label such as ``compressed protocol'' hides the family heterogeneity Appendix \ref{app:p0tables} records.

\hypertarget{app:negofail}{%
\section{Negotiation Failure}\label{app:negofail}}

Implemented C1 is \ttsplit{data/p1/dialogues\_c1\_nego.jsonl}. Runner-labeled phases are \texttt{handshake}, \texttt{schema\_nego}, \texttt{compressed}, and \texttt{english}. Source: \ttsplit{results/p2/c1\_negotiation.md}. No API calls.

The archived C1 in \ttsplit{data/p1/dialogues.jsonl} is the unimplemented contrast (injected canonical schema, turn-counter compression, zero handshake / schema\_nego turns). That corpus remains in \ttsplit{results/p2/c1\_phase\_breakdown.md} and is not deleted.

\end{multicols}
\begin{appendixwide}
\hypertarget{implemented-c1-phase-turns-headline-mid-n135}{%
\subsection{\texorpdfstring{Implemented C1 phase turns (headline \texttt{mid}, n=135)}{Implemented C1 phase turns (headline mid, n=135)}}\label{implemented-c1-phase-turns-headline-mid-n135}}

Mean / median / range of \textbf{dialogue-level} phase counts (A+B combined in the mean-turn column of \texttt{c1\_negotiation.md}):
\appendixwidetable
\centering
\footnotesize
\captionof{table}{Implemented C1 dialogue-level phase turns on the headline \texttt{mid} pair (n=135), as mean, median, and range.}\label{tab:a-phase-turns}
\begin{tabular}{@{}lrrr@{}}
\toprule
\textbf{Phase} & \textbf{mean turns} & \textbf{median} & \textbf{range} \\
\midrule
handshake & 2 & 2 & 2--2 \\
schema\_nego & 2.35 & 2 & 2--6 \\
compressed & 5.59 & 4 & 0--16 \\
english & 0.874 & 0 & 0--12 \\
\bottomrule
\end{tabular}
\end{appendixwide}
\begin{multicols}{2}

\ttsplit{schema\_agreed\_at\_turn} among agreed dialogues (all pairs, n=285): mean 3.4, median 3, range 3--7.

\end{multicols}
\begin{appendixwide}
\hypertarget{schema-agreement-not-canonical-confirmation}{%
\subsection{Schema agreement (not canonical confirmation)}\label{schema-agreement-not-canonical-confirmation}}

\texttt{schema\_agreed} is harness \texttt{SCHEMA\ ACCEPT} of a pending proposal. It is not \texttt{schema\_divergence} and not \texttt{CANONICAL\_SCHEMA}.
\appendixwidetable
\footnotesize
\captionof{table}{Implemented C1 \texttt{schema\_agreed} counts and rates by model pair, and pooled.}\label{tab:a-schema-agreed}
\begin{tabular}{@{}lrrr@{}}
\toprule
\textbf{Pair} & \textbf{n} & \textbf{schema\_agreed} & \textbf{rate} \\
\midrule
\ttsplit{mid} & 135 & 121 & 0.896 \\
\ttsplit{haiku-qwen} & 135 & 133 & 0.985 \\
\ttsplit{nano-nano} & 135 & 31 & 0.230 \\
All implemented C1 & 405 & 285 & 0.704 \\
\bottomrule
\end{tabular}
\end{appendixwide}
\begin{multicols}{2}

\ttsplit{schema\_matches\_canonical} is 0/285. Ten raw \texttt{schema\_text} samples: \ttsplit{results/p2/c1\_schema\_text\_samples.md}.

\texttt{schema\_nego\_failed} is 114/405. Those exits: turn\_cap 109, impasse 4, agreement 1. Six further dialogues are neither agreed nor failed; they ended the task without \texttt{SCHEMA\ ACCEPT}.

\hypertarget{handshake}{%
\subsection{Handshake}\label{handshake}}

\texttt{handshake\_pass} 85/405 = 0.210. Party A emits the harness six-word challenge in 405/405 dialogues. Failed B wires (raw): \ttsplit{results/p2/c1\_handshake\_fail\_wires.md}. Cross-tab: pass 79/85 schema-agreed; fail 206/320 schema-agreed.

\hypertarget{negotiation-interval-cost}{%
\subsection{Negotiation-interval cost}\label{negotiation-interval-cost}}

Implemented headline \(C_{\mathrm{neg}}\) at \texttt{h\ =\ 0}: A 0.000513, B 0.000816 (n=135). At \texttt{h\ =\ 0.9}: A 0.000185, B 0.000198. Unimplemented headline \(C_{\mathrm{neg}}\) was 0.

\hypertarget{refusal-silence-misunderstanding}{%
\subsection{Refusal, silence, misunderstanding}\label{refusal-silence-misunderstanding}}

The planned three-way split of handshake failures into refusal, silence, and misunderstanding was never logged, so there is no table here and no cell is filled with a guess. \texttt{handshake\_pass=False} is the logged verifier miss and it is not that split. The file the split would have come from, \ttsplit{results/appendix\_a/negotiation\_failures.jsonl}, was not collected. Successful schema adoption in the implemented overlay is Table \ref{tab:a-schema-agreed}.

\hypertarget{sensitivity-to-negotiation-failure-probability-not-produced}{%
\subsection{Sensitivity to negotiation failure probability (not produced)}\label{sensitivity-to-negotiation-failure-probability-not-produced}}

A party-specific break-even horizon over a varying analytic negotiation failure probability was not produced. Refusal, silence, and misunderstanding counts do not exist, and no \texttt{q} estimate is substituted.

\hypertarget{app:overlay}{%
\section{Secondary and Full-Precision Overlay Results}\label{app:overlay}}

Section 7 reports the headline \texttt{mid} overlay. This appendix carries the secondary tables it names, the compliance figure, the analysis plan's three-cell correspondence, and the same headline rows at full precision. Sources are the analysis scripts in Appendix \ref{app:artifacts}.

\end{multicols}
\begin{appendixwide}
\hypertarget{search-failure-wide-and-narrow-impasse}{%
\subsection{Search failure: wide and narrow impasse}\label{search-failure-wide-and-narrow-impasse}}

Empty-F impasse is the correct conclusion. Wide/narrow impasse is \texttt{search\_failure}.
\appendixwidetable
\footnotesize
\captionof{table}{Search-failure (wide/narrow impasse) rate by condition, for the headline \texttt{mid} pair and pooled over all three pairs.}\label{tab:searchfail}
\begin{tabular}{@{}lrrrr@{}}
\toprule
\textbf{Scope} & \textbf{C0} & \textbf{C1 implemented} & \textbf{C3} & \textbf{C1 unimplemented} \\
\midrule
Headline \ttsplit{mid} & 39/77 = 0.506 & 70/90 = 0.778 & 61/75 = 0.813 & 70/76 = 0.921 \\
All three pairs & 102/241 = 0.423 & 136/270 = 0.504 & 116/231 = 0.502 & 161/231 = 0.697 \\
\bottomrule
\end{tabular}
\end{appendixwide}
\begin{multicols}{2}

C0 and C3 headline rates are 0.506 and 0.813, unchanged from the archived analysis. P1 has no registered gate, so these are reported values rather than preregistered ones.

\end{multicols}
\begin{appendixwide}
\appendixwidetable
\footnotesize
\captionof{table}{Majority exit and turn-cap share by pair, C0 and C3 of the archived overlay.}\label{tab:pairexit}
\begin{tabular}{@{}llrlr@{}}
\toprule
\textbf{Pair} & \textbf{C0 majority exit} & \textbf{C0 turn\_cap} & \textbf{C3 majority exit} & \textbf{C3 turn\_cap} \\
\midrule
\ttsplit{mid} & impasse & 52/118 & impasse & 20/112 \\
\ttsplit{haiku-qwen} & impasse & 0/120 & impasse & 0/118 \\
\ttsplit{nano-nano} & turn\_cap & 120/122 & turn\_cap & 106/119 \\
\bottomrule
\end{tabular}
\end{appendixwide}
\begin{multicols}{2}

\end{multicols}
\begin{appendixwide}
\hypertarget{outcome-class-archived-overlay}{%
\subsection{Outcome class (archived overlay)}\label{outcome-class-archived-overlay}}
\appendixwidetable
\footnotesize
\captionof{table}{Outcome class for the archived overlay. Equivalence is the \texttt{error\_*} columns only. C1 counts here are the unimplemented corpus. Do not rank conditions by \texttt{correct}.}\label{tab:outcome}
\begin{tabular}{@{}lllllr@{}}
\toprule
\textbf{Condition} & \textbf{correct} & \textbf{search\_failure} & \textbf{error\_empty} & \textbf{error\_below\_reservation} & \textbf{n} \\
\midrule
C0 & 80 & 274 & 3 & 3 & 360 \\
C1 unimplemented & 116 & 209 & 8 & 11 & 344 \\
C3 & 94 & 243 & 4 & 8 & 349 \\
C2 / C4 / C5 & not measured & not measured & not measured & not measured & 0 \\
\bottomrule
\end{tabular}
\end{appendixwide}
\begin{multicols}{2}

Implemented C1 outcome-class tallies are not substituted into Table \ref{tab:outcome}. Headline implemented C1 \texttt{correct} is 0.281 {[}0.207, 0.356{]} (n=135).

\end{multicols}
\begin{appendixwide}
\hypertarget{compressed-wire-compliance-against-cost}{%
\subsection{Compressed-wire compliance against cost}\label{compressed-wire-compliance-against-cost}}
\appendixwidetable
\includegraphics[width=0.92\textwidth]{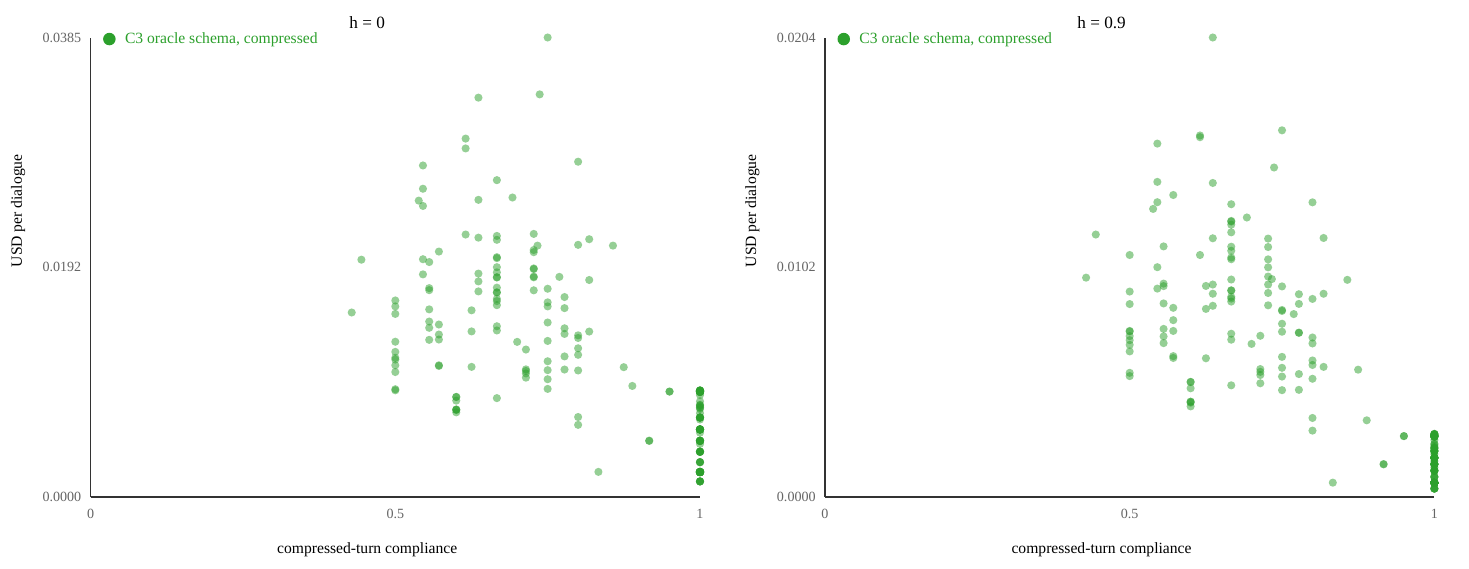}
\captionof{figure}{Compressed-turn compliance against party-sum listed cost, one mark per C3 dialogue on cross-vendor pairs of the archived overlay. Compliance here is the \textbf{mean over that dialogue's compressed turns}. C0 and implemented C1 are absent.}\label{fig:compliance}
\end{appendixwide}
\begin{multicols}{2} C3 is the whole of Figure \ref{fig:compliance}: it is the only measured condition, with mean 0.825 and a minimum of 0.429 over 230 dialogues. C0 has no compressed turn to check and implemented C1 negotiates its grammar per dialogue, so neither records a violation rate and plotting their default 1.0 would show the instrument rather than the protocol. Section 9 and I.7 use a different aggregation of the same field, the \textbf{minimum of the two party rates}, and the two are not interchangeable.

\end{multicols}
\begin{appendixwide}
\hypertarget{the-analysis-plans-listed-cost-cell}{%
\subsection{The analysis plan's listed-cost cell}\label{the-analysis-plans-listed-cost-cell}}

The analysis plan fixes a three-cell correspondence on \textbf{point means of total listed cost}:
\appendixwidetable
\scriptsize
\captionof{table}{The analysis plan's three-cell correspondence on point means of total listed cost, with what the implemented overlay shows for each cell.}\label{tab:planmap}
\begin{tabularx}{\textwidth}{@{}ll>{\hsize=0.845\hsize\RaggedRight\arraybackslash}X>{\hsize=1.155\hsize\RaggedRight\arraybackslash}X@{}}
\toprule
\textbf{C3 vs C0} & \textbf{C1 vs C0} & \textbf{Plan sentence} & \textbf{Implemented overlay} \\
\midrule
C3 \textless{} C0 & C1 \textless{} C0 & Compression helps and negotiation is recouped & \textbf{True for listed totals} (C1 0.00379 \textless{} C0 0.00735) and \textbf{false for interpretive \(N^*\)} (Section 8) \\
C3 \textless{} C0 & C1 ≥ C0 & Compression helps but negotiation eats it. Share the schema up front & \textbf{Rejected} on totals \\
C3 ≥ C0 & any & Compression itself is a loss, with or without a switch & \textbf{Rejected}: C3 0.00499 \textless{} C0 0.00735 \\
\bottomrule
\end{tabularx}
\end{appendixwide}
\begin{multicols}{2}

The selected listed-cost cell is not the result. Headline \texttt{schema\_agreed} is 0.896 while task \texttt{agreement} is 0.0667. \(N^*\) exceeds measured C0 length at every tested \texttt{h}. Compression of a failed bargain is still cheaper on the clock; that is not recouped protocol negotiation.

\end{multicols}
\begin{appendixwide}
\hypertarget{full-precision}{%
\subsection{Full precision}\label{full-precision}}

Table \ref{tab:headline} rounds to three significant figures. The same rows, with 95\% percentile intervals over 10,000 bootstrap replicates:
\appendixwidetable
\scriptsize
\captionof{table}{Full-precision headline \texttt{mid} overlay with 95\% percentile intervals.}\label{tab:headline-full}
\begin{tabularx}{\textwidth}{@{}>{\hsize=1.210\hsize\RaggedRight\arraybackslash}X>{\hsize=1.171\hsize\RaggedRight\arraybackslash}X>{\hsize=0.780\hsize\RaggedRight\arraybackslash}X>{\hsize=0.780\hsize\RaggedRight\arraybackslash}X>{\hsize=1.132\hsize\RaggedRight\arraybackslash}X>{\hsize=1.132\hsize\RaggedRight\arraybackslash}X>{\hsize=0.663\hsize\RaggedRight\arraybackslash}X>{\hsize=1.132\hsize\RaggedRight\arraybackslash}X@{}}
\toprule
\textbf{Condition} & \textbf{agreement} & \textbf{impasse} & \textbf{turn\_cap} & \textbf{Total h=0} & \textbf{Total h=0.9} & \textbf{Turns} & \textbf{Cost / turn} \\
\midrule
C0 English & 0.00847 {[}0, 0.0254{]} (n=118) & 0.551 {[}0.458, 0.644{]} & 0.441 {[}0.347, 0.534{]} & 0.007347 {[}0.007009, 0.007656{]} & 0.002280 {[}0.002173, 0.002379{]} & 17.6 {[}16.9, 18.3{]} & 0.000411 {[}0.000406, 0.000416{]} \\
C1 implemented handshake+schema & 0.0667 {[}0.0296, 0.111{]} (n=135) & 0.785 {[}0.719, 0.852{]} & 0.148 {[}0.089, 0.207{]} & 0.003790 {[}0.003442, 0.004153{]} & 0.001115 {[}0.001000, 0.001235{]} & 10.8 {[}10.0, 11.6{]} & 0.000335 {[}0.000328, 0.000343{]} \\
C3 oracle schema, compressed & 0 {[}0, 0{]} (n=112) & 0.821 {[}0.750, 0.893{]} & 0.179 {[}0.107, 0.250{]} & 0.004994 {[}0.004490, 0.005513{]} & 0.001518 {[}0.001372, 0.001672{]} & 12.3 {[}11.3, 13.3{]} & 0.000389 {[}0.000382, 0.000396{]} \\
\bottomrule
\end{tabularx}
\end{appendixwide}
\begin{multicols}{2}

The matched contrasts at full precision, headline \texttt{mid}, \texttt{h\ =\ 0}: implemented C1 − C0 total −0.003551 {[}−0.004038, −0.003021{]} (n=118), C1 − C3 total −0.001220 {[}−0.001885, −0.000558{]} (n=112), and C3 − C0 total −0.002369 {[}−0.002981, −0.001748{]} (n=99). Values are unchanged from the analysis record in Appendix \ref{app:artifacts}.

\hypertarget{the-superseded-negotiation-cost-numerator}{%
\subsection{The superseded negotiation-cost numerator}\label{the-superseded-negotiation-cost-numerator}}

An earlier draft set \(C_{\mathrm{neg}}\) to C1 − C3. On the headline pair that contrast is negative (at \texttt{h\ =\ 0}, −0.00108 for A and −0.00192 for B; matched n=95 on the unimplemented corpus). Unimplemented C1 mean length is 5.15 turns and C3 is 12.3, so \texttt{C1\ −\ C3} mixes negotiation cost with a length difference. Negative overhead is not a possible handshake cost. This paper records that definition error. The corrected numerator is handshake plus schema-negotiation phase cost. On the unimplemented corpus that numerator was 0. On the implemented overlay it is positive and yields \(N^*\) above the measured C0 length. The paper does not report the superseded negative quantity as \(N^*\).

\hypertarget{compliance-and-impasse-in-the-pooled-c3-data}{%
\subsection{Compliance and impasse in the pooled C3 data}\label{compliance-and-impasse-in-the-pooled-c3-data}}

Low compliance is associated with impasse in the pooled C3 data. This paragraph aggregates the field as the \textbf{minimum of the two party rates}, which is stricter than the per-dialogue mean that Figure \ref{fig:compliance} plots, and pools all three pairs. Dialogues below 0.5 end in impasse in 53 of 72 cases. Dialogues at or above 0.9 do so in 93 of 212 cases. This comparison does not identify a cause.

The two rates in that comparison are drawn from different pairs, which is what makes the pooled contrast weaker than it looks. The 212 high-compliance dialogues are 109 \texttt{mid} and 103 \texttt{nano-nano}, and no \texttt{haiku-qwen} dialogue is among them: that pair's party-A rate keeps its minimum below the threshold. The 4 impasses outside \texttt{mid} therefore come from self-play, whose C3 exit is turn-cap in 97 of its 103 high-compliance dialogues. Pooling a cross-vendor pair that mostly deadlocks with a self-play pair that mostly runs out of turns produces the 93 of 212, not a compliance effect.

The low-compliance cell is concentrated in one pair. Haiku-qwen C3 has mean A compliance 0.375 and B compliance 1.000. That pair concentration confounds format compliance with model composition. Failed compressed syntax might block bargaining expressions, but these data cannot isolate that mechanism.

\end{multicols}
\begin{appendixwide}
\hypertarget{break-even-against-the-cache-hit-sweep}{%
\subsection{Break-even against the cache-hit sweep}\label{break-even-against-the-cache-hit-sweep}}

Table \ref{tab:nstar} in Section 8.2 gives these values as numbers.
\appendixwidetable
\includegraphics[width=0.92\textwidth]{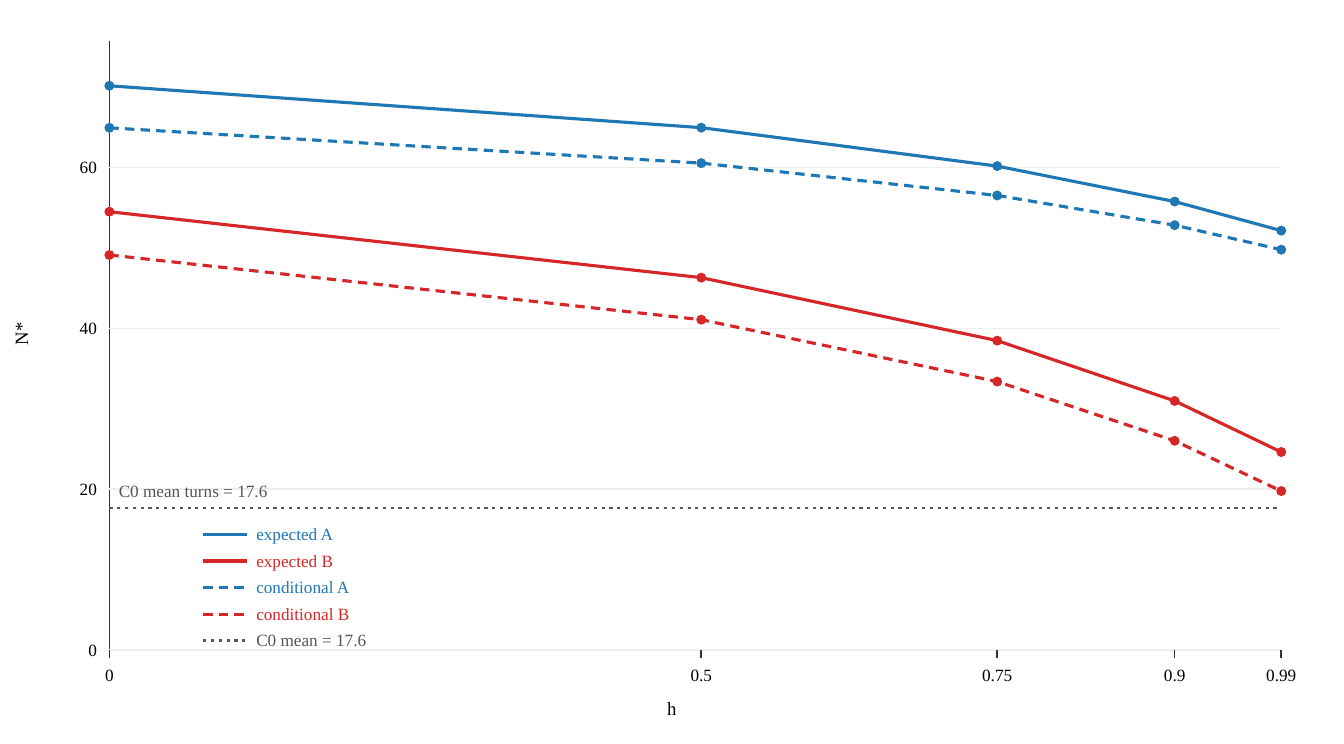}
\captionof{figure}{Interpretive \(N^*\) versus cache-hit rate \texttt{h} for headline implemented C1. Solid curves are expected (all dialogues); dashed curves are conditional on \texttt{schema\_agreed}. Party A and B are separate. The horizontal line is the measured C0 mean of 17.6 turns. \(C_{\mathrm{neg}}\) is handshake plus schema-negotiation phase cost.}\label{fig:breakeven}
\end{appendixwide}
\begin{multicols}{2}

\end{multicols}
\begin{appendixwide}
\hypertarget{app:artifacts}{%
\section{Artifact Index}\label{app:artifacts}}

Every number, table, and figure in the body is generated from append-only records under \texttt{data/**} and \texttt{results/**} and from the scripts under \texttt{scripts/**} and \texttt{harness/**}. The body prose cites those artifacts through this index rather than inline. No file listed here is recomputed, regenerated, or modified when the manuscript is assembled. The two implemented- and unimplemented-C1 dialogue files are cited inline in the body as well, because the contrast between them is the subject of several sentences.
\appendixwidetable
\scriptsize
\captionof{table}{Artifact index: the append-only record or script behind each body claim, table, and figure.}\label{tab:artifacts}
\begin{tabularx}{\textwidth}{@{}>{\hsize=0.741\hsize\RaggedRight\arraybackslash}X>{\hsize=0.828\hsize\RaggedRight\arraybackslash}X>{\hsize=1.431\hsize\RaggedRight\arraybackslash}X@{}}
\toprule
\textbf{Section} & \textbf{Claim / table} & \textbf{Source path} \\
\midrule
§4.1 & cache-price and TTL sweep parameters & \ttsplit{data/design/cost\_parameters.json} \\
§5.1 & difficulty band feasible-set fraction diagnostic & \ttsplit{results/negoenv/difficulty-diagnostic.md} \\
§5.1 & frozen overlay pair inventory & \ttsplit{data/design/model\_pairs\_core.json} \\
§5.1 & C3 self-play compliance screen & \ttsplit{results/p1/model\_screening.md} \\
§5.1 & shuffle manifest (seed 20260824) & \ttsplit{data/p1/run\_manifest\_core.json} \\
§5.2 & condition interfaces & \ttsplit{harness/conditions.py} \\
§5.2 & condition prompts and permissions & \ttsplit{harness/runner.py} \\
§5.2, §7 & archived C0/C3 and unimplemented C1 & \ttsplit{data/p1/dialogues.jsonl} (also inline) \\
§5.2, §7 & implemented C1 renegotiation overlay & \ttsplit{data/p1/dialogues\_c1\_nego.jsonl} (also inline) \\
Appendix \ref{app:p0gate} & P0 items & \ttsplit{data/p0/items.jsonl} \\
§6.2 & P0b preregistration & \ttsplit{PREREGISTRATION-p0b.md} \\
§6.2 & P0b items & \ttsplit{data/p0b/items.jsonl} \\
§6.3 & P0b verdict & \ttsplit{results/p0b/VERDICT.md} \\
§6.3 & P0b raw usage & \ttsplit{data/p0b/raw\_usage.jsonl} \\
§6.3 & P0b verdict generator & \ttsplit{scripts/p0b\_verdict.py} \\
§6.4 & P0b confirmation-half family estimates & \ttsplit{results/p0b/families.md} \\
§6.4 & digit-versus-symbol post-hoc & \ttsplit{results/p0b/POSTHOC-digit-vs-symbol.md} \\
§7 intro & archived C0/C3 analysis & \ttsplit{scripts/p2\_analyze.py} \\
§7 intro & implemented C1 analysis & \ttsplit{scripts/p2\_c1\_nego\_analyze.py} \\
§7 Table \ref{tab:headline} and overhead contrast & implemented C1 party-separated cells & \ttsplit{results/p2/core\_answer\_v2.md} \\
§7 Table \ref{tab:headline} & archived C0/C3 cells & \ttsplit{results/p2/core\_answer.md} \\
§7 protocol gate & schema-agreement rates & \ttsplit{results/p2/c1\_negotiation.md} \\
Appendix \ref{app:overlay}, Table \ref{tab:outcome} & archived outcome-class tallies & \ttsplit{results/p2/table3\_outcome\_class.md} \\
§8, Figure \ref{fig:breakeven} & break-even versus cache-hit rate & \ttsplit{results/p2/fig1\_breakeven.svg} from \ttsplit{results/p2/c1\_nego\_breakeven.csv} \\
§7, Figure \ref{fig:cheaper} & cheaper for whom & \ttsplit{results/p2/fig2\_cheaper\_for\_whom.svg} \\
Appendix \ref{app:overlay}, Figure \ref{fig:compliance} & compliance versus cost & \ttsplit{results/p2/fig3\_compliance\_cost.svg} \\
§8.1 & C3/C0 per-turn ratio \ttsplit{d} & \ttsplit{results/p2/interpretive\_breakeven.csv} \\
§8.1 & vendor prices & \ttsplit{data/design/cost\_parameters.json} \\
§8.1 & negotiation phase costs & \ttsplit{results/p2/c1\_negotiation.md} \\
§8.1 & full-precision break-even & \ttsplit{results/p2/c1\_nego\_breakeven.csv} \\
Appendix \ref{app:overlay} & unimplemented \(C_{\mathrm{neg}} = 0\) record & \ttsplit{results/p2/interpretive\_breakeven.csv} \\
§9, Limitations & archived 6x6 shuffled remainder & \ttsplit{data/p1/partial\_full\_matrix\_20260824/} \\
Limitations & aborted first P1 main run & \ttsplit{data/p1/aborted\_run\_20260824/} \\
Limitations & C1 compliance-check skip & \ttsplit{harness/runner.py} \\
\bottomrule
\end{tabularx}
\end{appendixwide}
\begin{multicols}{2}

\closeappendixcolumns

\clearpage

\end{document}